\documentclass[10pt,twocolumn,letterpaper]{article}

\usepackage[pagenumbers]{cvpr} %

\usepackage{xspace}

\usepackage[super]{nth}

\newcommand{\methodname}{PixVid\xspace}

\usepackage{bm}
\usepackage{multirow}  %
\usepackage{colortbl}  %
\usepackage{verbatim}

\makeatletter
\renewcommand\paragraph{\@startsection{paragraph}{4}{\z@}%
  {1ex \@plus 0.2ex \@minus 0.2ex}%
  {-1em}%
  {\normalfont\normalsize\bfseries}}
\makeatother

\DeclareMathOperator*{\argmin}{arg\,min}

\definecolor{cvprblue}{rgb}{0.21,0.49,0.74}
\usepackage[pagebackref,breaklinks,colorlinks,allcolors=cvprblue]{hyperref}
\usepackage{multirow}
\usepackage{float}

\def\confName{CVPR}
\def\confYear{2027}

\title{Back to Pixels: Lifting Latent Video Diffusion into Pixel Space}

\author{
Yanke Song$^{1}$\quad
Shuo Chen$^{1,2}$\quad
Chris Rockwell$^{1}$\quad
Guande He$^{1}$\quad
Nan Liu$^{1}$\quad
Desai Xie$^{1}$\quad
Tianpei Gu$^{1}$\\
Yong-Hyun Park$^{1,3}$\quad
Aryan V S$^{1}$\quad
Dennis Yuren Menn$^{1,4}$\quad
Ming-Hsuan Yang$^{2}$\quad
Xun Huang$^{1}$\\[0.5em]
{\normalsize
$^{1}$Roblox\qquad
$^{2}$University of California, Merced\qquad
$^{3}$University of Pennsylvania\qquad
$^{4}$University of Texas at Austin}
}

\begin{document}
\twocolumn[{%
\renewcommand\twocolumn[1][]{#1}%
\maketitle
\begin{center}
  \centering
  \includegraphics[width=\textwidth]{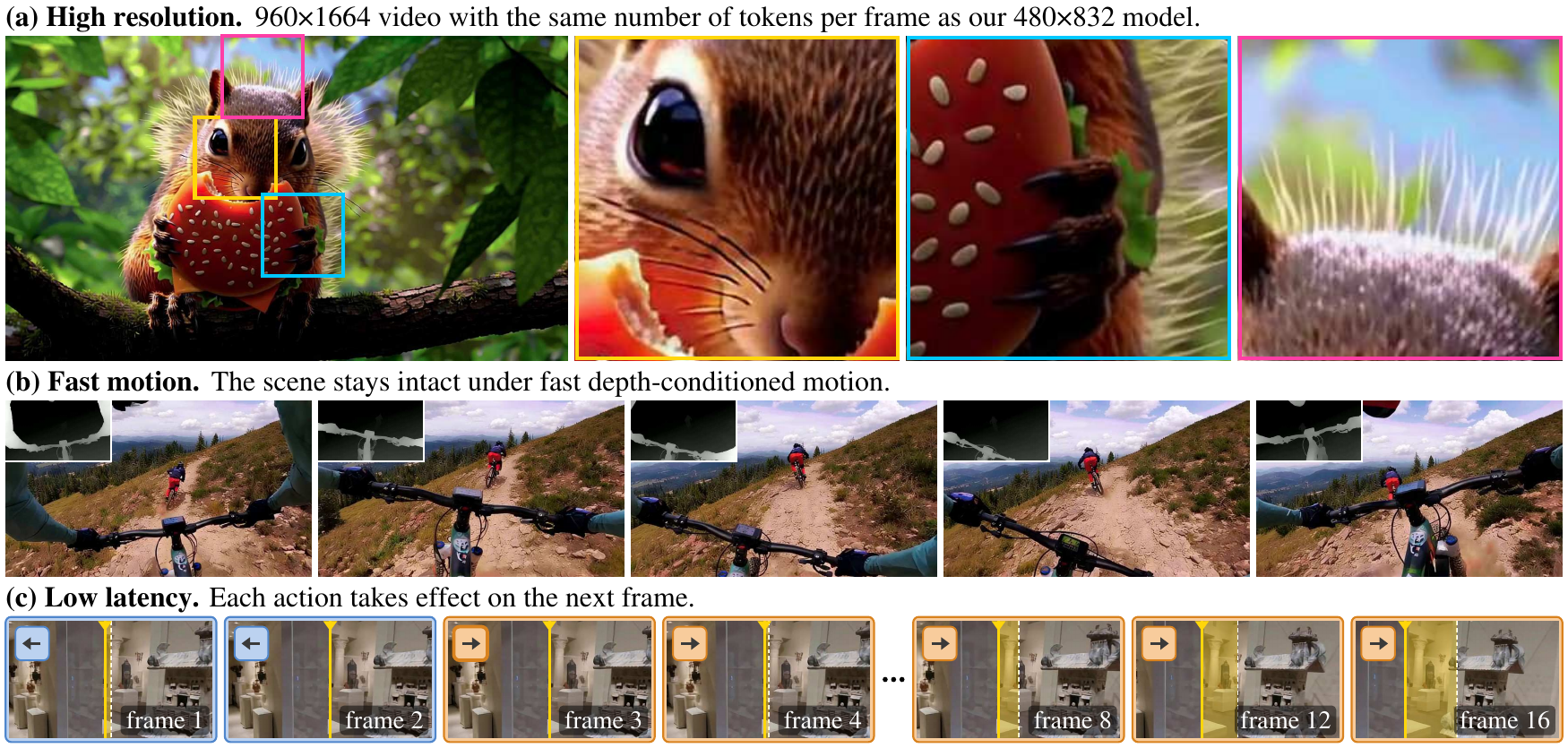}
  \captionof{figure}{%
    \methodname lifts a pretrained latent video model into pixel space, enabling:
    \textbf{(a)} scaling to $960\times1664$ video at minimal extra cost per frame over $480\times832$, at or above the pretrained latent model's quality;
    \textbf{(b)} better motion handling in video editing than its latent counterpart; and
    \textbf{(c)} response to each action at the next frame, compared with latent models which cannot respond until the next block of frames.
    The dashed line marks the frame center, the yellow line a fixed scene point, and the shading how far the view has turned.
    }
  \label{fig:teaser}
\end{center}
\vspace{0.5em}
}]

\begin{abstract}

Frontier video generation models work in a VAE latent space that compresses video in both space and time.
This compression made training tractable, but now limits high-resolution, controllable, and interactive generation -- doubling resolution quadruples token count, conditions apply to groups of frames rather than individual ones, and each response must wait for the rest of its multi-frame block.
We present \methodname, a simple yet effective method that lifts a pretrained latent video model to directly generate every frame in pixel space.
\methodname keeps the diffusion transformer and replaces the VAE decoder with a convolutional head that renders its outputs into pixels.
We find that predicting clean video rather than velocity, together with a larger noise shift, is key to quality.
Finetuned from Wan2.1-14B, \methodname is competitive with the latent baseline within $\sim 1.3$M training clips.
Working framewise in pixels, it generates at $2\times$ higher resolution with the same per-frame token count, keeps depth-conditioned video editing aligned under fast motion, and responds to each action at the next frame. Project page: \url{https://chenshuo20.github.io/PixVid/}

\end{abstract}
    
\section{Introduction}
\label{sec:intro}

Early video diffusion models~\cite{ho2022video, ho2022imagen} generated videos in their native form, as sequences of pixel frames.
As models scaled up, the field made two shifts away from pixel space:
The first followed latent diffusion~\cite{rombach2022high}, moving generation into the latent space of a pretrained variational autoencoder (VAE)~\cite{kingma2014auto}, initially with one latent per frame~\cite{blattmann2023align, blattmann2023stable}.
The second extended the compression to time, using 3D VAEs to merge every few frames into a single latent that corresponds to no individual frame~\cite{yang2025cogvideox, wan2025wan}.
For generating short clips offline at moderate resolution, this was a favorable trade: compression made training tractable at scale, at the cost of a frozen VAE that ruled out end-to-end training and capped quality at its reconstruction~\cite{chen2025dc, esser2024scaling, skorokhodov2025improving}.

Video generation models have since outgrown this setting: they now power real-time interactive world models~\cite{he2025matrix, sun2025worldplay, hong2025relic} through causal few-step distillation~\cite{yin2025slow, huang2026self}, and serve as geometry-conditioned renderers for editing and games~\cite{vace, nvidia2025cosmostransfer1, nvidia2025cosmos}.
Yet these systems inherit the VAE’s fixed compression in space and time, which works against each new demand.
Under a real-time compute budget, doubling the resolution yields $4\times$ more tokens and $16\times$ more attention compute, while cutting tokens requires a more aggressive VAE that further reduces quality.
Precise editing requires conditions aligned to individual frames, but when several frames share one latent, fast motion blurs and objects deform (Fig.~\ref{fig:d2v}).
Temporal compression also limits responsiveness. Since common latent world models~\cite{yin2025slow, huang2026self, yang2025longlive, he2025matrix, sun2025worldplay, huang2026solarwmopendatascalable} autoregressively generate frames in blocks, the frame corresponding to a control cannot be generated until all controls for the full block are received. This results in a minimum control-to-screen latency proportional to the block size, a barrier that no model-side optimization can remove.

These limitations motivate removing the VAE altogether and generating every frame directly as pixels, which naively would bring back the token counts and training costs that latent models avoid.
Recent pixel-space image models overcome both challenges~\cite{li2026back, yu2026pixeldit, chen2026dip, chen2026l2p, jiang2026empirical, chen2026asymmetric} by modeling semantics over large pixel patches~\cite{li2026back} and delegating fine details to a lightweight decoder~\cite{chen2026dip}, thus keeping the token count and training cost reasonable. 
Moreover, although training video models from scratch demands prohibitive data and compute resources, a pretrained latent model already captures high-level semantics and dynamics. Hence, finetuning it into a pixel space video model only requires changing the output interface, which is a more feasible task. 

\begin{figure*}[!t]
  \centering
  \includegraphics[width=\textwidth]{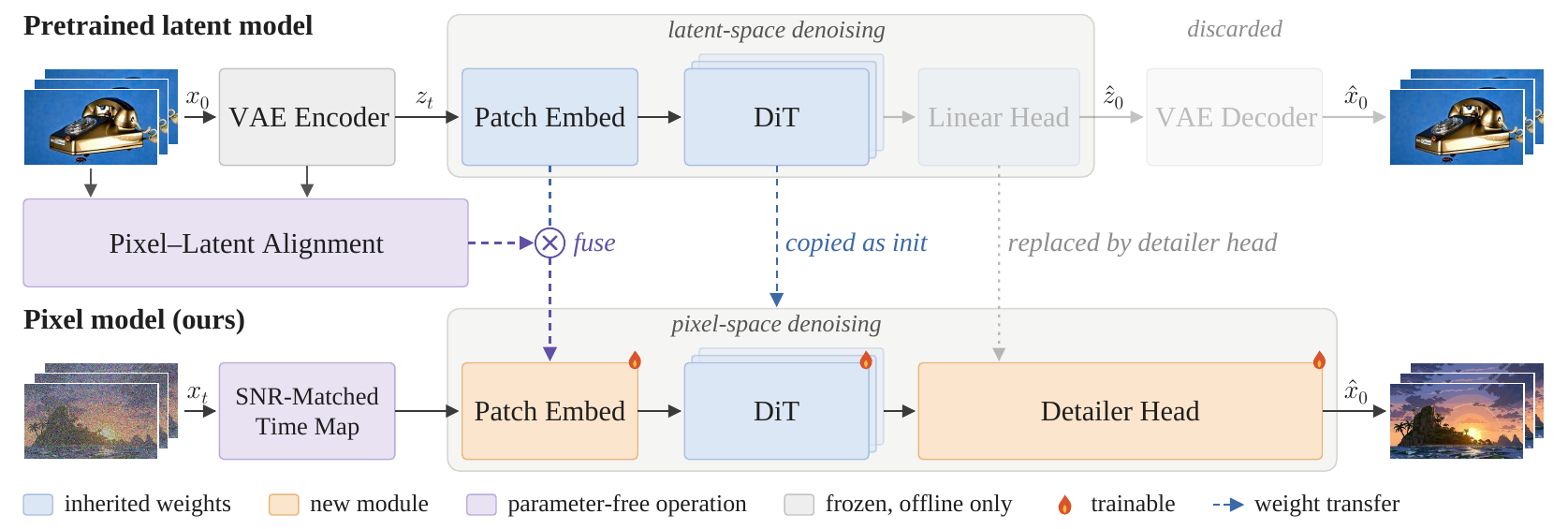}
  \caption{\textbf{Lifting a pretrained latent video DiT to pixel space.}
  \emph{Top:} the pretrained latent model. 
  A linear pixel-latent alignment, fitted offline, is fused with the pretrained patch embedding to obtain a pixel patch embedding.
  \emph{Bottom:} PixVid. 
  A timestep calibration rescales noisy pixel input to the signal-to-noise ratio the pretrained DiT was trained on. 
  The DiT is initialized from the latent model, and a convolutional detailer head replaces the linear head and VAE decoder to predict the clean video directly.}
  \label{fig:method}
\end{figure*}

In this work, we present \textbf{\methodname}, a recipe that lifts a pretrained latent video diffusion model into a \emph{frame-wise pixel-space} model, in which every token covers a $1\times p\times p$ patch of exactly one frame, so no condition or action spans multiple frames.
\methodname\ keeps the diffusion transformer (DiT) and replaces both the encoder and the decoder from the VAE (Fig.~\ref{fig:method}).
At input, a patch embedding that initializes from the VAE encoder allows the
DiT to start training from close to its pretrained latent distribution.
At output, a lightweight detailer head renders the DiT tokens into pixels; it is convolutional to suppress the patch-grid artifacts of a linear head~\cite{wei2026pixelumm}, and causal in time to allow autoregressive generation.
We further identify the following key techniques: 1) predicting clean video ($x$-prediction) rather than velocity, since velocity prediction forces the network to model full-rank per-pixel noise~\cite{li2026back}; 2) using a larger noise shift, since redundant pixels retain more signal at the same noise level; and 3) using the Euler sampler, since pixel color values are numerically sensitive to higher order solvers like UniPC~\cite{zhao2023unipc}.

Finetuned from Wan2.1-14B~\cite{wan2025wan}, \methodname\ converges within $1.3$M training clips sampled from $64$K videos, or about $4$K B200 GPU-hours. It achieves competitive quality with the original latent model on VBench~\cite{huang2024vbench}, and the recipe transfers across Wan2.1 and Wan2.2-based models with different VAEs.
Generating frame by frame in pixels, it introduces several benefits, as shown in \cref{fig:teaser}. 
\emph{Resolution:} doubling the patch size lets \methodname generate $960\times1664$ with the same number of tokens per frame as at $480\times832$.
\emph{Control:} with the same control branch as the latent model, depth-conditioned editing keeps $98\%$ of the source's fast motion, versus $83\%$ for an identically finetuned latent model.
\emph{Latency:} after autoregressive distillation, each camera control takes effect at the next frame rather than the next block, cutting median latency from $495$ ms to $34$ ms while still streaming in real time.

To summarize, our main contributions are as follows:
\begin{itemize}
    \item We propose \textbf{\methodname}, a recipe that lifts a pretrained latent video diffusion model into a frame-wise pixel-space model, converging quickly to competitive quality.
    \item We identify the design choices that make this lift work: a causal convolutional detailer head with $9\times$ fewer patch seams than a linear head, $x$-prediction with $80\times$ less pixel noise than $v$-prediction, and a larger noise shift than the latent model.
    \item We show frame-wise pixel generation overcomes three structural limitations of latent models: resolution scaling, control alignment under fast motion, and control-to-screen latency.
\end{itemize}

\section{Related Work}
\label{sec:related}

\noindent \textbf{Pixel-Space Diffusion.} Latent diffusion~\cite{rombach2022high, peebles2023scalable, blattmann2023align} has been favored for efficient training using spatial and temporal compression for video~\cite{yang2025cogvideox, wan2025wan, videoworldsimulators2024}, at the price of a VAE reconstruction bottleneck. Pixel-space diffusion models for images~\cite{ho2020denoising, song2020score, dhariwal2021diffusion, karras2022elucidating} and videos~\cite{ho2022video, ho2022imagen, singer2022make, bar2024lumiere} bypass a VAE but become costly at high resolution, historically requiring cascaded or multi-resolution designs~\cite{ho2022cascaded, saharia2022photorealistic, ramesh2022hierarchical, gu2024matryoshka, crowson2024scalable, teng2024relay}. Recent methods improved pixel-space image generation, training from scratch~\cite{hoogeboom2023simple, hoogeboom2025simpler, chen2025pixelflow, li2026back, yu2026pixeldit, chen2026dip, ma2026deco, wang2026pixnerd, liu2026there, lu2026one, chen2026asymmetric, mukhopadhyay2026rethinking} or finetuning from latent checkpoints to make training tenable~\cite{chen2026l2p, jiang2026empirical, chen2026asymmetric}. 
Pixel-space video generation is less explored. Concurrent works train from scratch, and they either focus on smaller class-conditioned benchmarks~\cite{vallaeys2026ujit} or have undesirable patch artifacts~\cite{wei2026pixelumm}. We finetune a pretrained latent video model, inheriting its prior capabilities and maintaining competitive quality.

\vspace{2.5pt}\noindent \textbf{Controllable Video Generation and Editing.} Early video editors applied image diffusion models with heuristics such as cross-frame shared self-attention~\cite{wu2023tune, qi2023fatezero, khachatryan2023text2video, geyer2024tokenflow} or propagated edits~\cite{bar2022text2live, yang2023rerender, ouyang2024codef}. Video diffusion models improve consistency and robustness, with recent models producing precise, aligned edits and structure-guided generation~\cite{vace, ye2025unic, ju2025editverse, nvidia2025cosmostransfer1, flux3, aleph2, seedance25, omni}. 

\vspace{2.5pt}\noindent \textbf{Interactive Video Generation.} Interactive generation produces frames as a stream, typically with autoregressive video diffusion models~\cite{chen2024diffusion, deng2025autoregressive, xie2025progressive, teng2025magi, chen2025skyreels}. Combined with few-step distillation~\cite{yin2024one, yin2024improved, yin2025slow, huang2026self, yang2025longlive, zhu2026causal, chen2026context}, latent world models respond to user actions across diverse scenes in real time~\cite{he2025matrix, mao2025yume, sun2025worldplay, hong2025relic}, 
but temporal compression delays each action, \eg by up to three frames for Wan's four-frame latents~\cite{wan2025wan}.
Per-frame latent~\cite{bruce2024genie, valevski2025diffusion} and pixel-space~\cite{alonso2024diffusion, dlss5} world models avoid this delay but are trained from scratch on a single genre. \methodname\ is finetuned from a general pretrained model and responds to actions at the next frame.

\section{Methods}
\label{sec:methods}

Our goal is to generate every frame directly in pixels without training a pixel-space video model from scratch, while removing the VAE bottleneck.
\methodname\ achieves this by lifting a pretrained latent video diffusion model into a frame-wise pixel-space model, as shown in Fig.~\ref{fig:method}.
We describe its architecture in Sec.~\ref{sec:architecture} and how we train it from the latent model in Sec.~\ref{sec:finetuning}, including key design choices.
We then adapt it to efficient high-resolution generation in Sec.~\ref{sec:high_resolution}, precise video editing in Sec.~\ref{sec:video_editing}, and low-latency autoregressive generation in Sec.~\ref{sec:autoregressive_gen}.

\subsection{Architecture}\label{sec:architecture}

\methodname\ keeps the DiT~\cite{peebles2023scalable} backbone of a pretrained latent video model and replaces the VAE: a linear patch embedding maps pixels to tokens, and a lightweight convolutional detailer head renders the DiT outputs directly into pixels.

\vspace{2.5pt}\noindent\textbf{Frame-Wise Tokenization.}
Latent video models merge several frames into one latent~\cite{yang2025cogvideox, wan2025wan}, so conditions apply to groups of frames.
\methodname\ instead gives each frame its own tokens, each covering a $1\times p\times p$ pixel patch.
Conditions then align with individual frames, and an autoregressive model can respond to actions at the next frame.
Although this approach uses more tokens per clip than with temporal compression, the spatial token count does not grow with resolution as it is determined by the adjustable patch size rather than a frozen VAE.
 
\vspace{2.5pt}\noindent\textbf{Detailer Head.}
Since the DiT captures global structure and leaves only local details to render, we implement the detailer as a lightweight 3D convolutional decoder, sharing this insight with pixel-space image models~\cite{chen2026dip, chen2026l2p, jiang2026empirical}.
Unlike their U-Net detailers, which also encode the noisy pixels of each patch, ours keeps only the decoder path and reads only the DiT tokens.
It stacks $L=\log_2 p$ up blocks, each upsampling by $1\times2\times2$, so the head maps each token to a $1\times p\times p$ pixel patch and predicts the clean video.
Its convolutions also run across neighboring tokens rather than within each patch, which largely suppresses checkerboard artifacts at token boundaries.
They are causal in time, so the detailer models local temporal structure without reading future frames and supports frame-by-frame generation.
See App.~\ref{sec:detailer} for details.

\subsection{Training from a Latent Model}\label{sec:finetuning}

\methodname finetunes end-to-end in pixel space from a pretrained latent model, with initialized DiT and input patchification, $\bm{x}$-prediction, a larger noise shift, and auxiliary losses.

\vspace{2.5pt}\noindent\textbf{Initialization.}
To retain the pretrained latent model's prior, we inherit its DiT weights and randomly initialize the detailer head. 
Following AsymFlow~\cite{chen2026asymmetric}, we approximate the VAE encoder with a linear map $\bm{A}$ fitted offline between pixel patches and their latents, and compose it with the pretrained patch embedding to initialize the pixel patch embedding.
We further employ an internal timestep calibration, with scale $c$ estimated alongside $\bm{A}$, to ensure the signal-to-noise ratios of latent and pixel spaces are aligned. The DiT thus initially receives inputs close to the latents it was trained on. Full details appear in App.~\ref{sec:patch_embed_init}.

\vspace{2.5pt}\noindent\textbf{$\bm{x}$-Prediction.}
Under flow matching~\cite{lipman2022flow}, a clean video $\bm{x}$ is corrupted with Gaussian noise $\bm{\epsilon}\sim\mathcal{N}(\bm{0},\bm{I})$ at noise level $\sigma\in[0,1]$ as $\bm{x}_\sigma=(1-\sigma)\,\bm{x}+\sigma\,\bm{\epsilon}$.
Following JiT~\cite{li2026back}, the network predicts the clean video $\hat{\bm{x}}=\bm{x}_\theta(\bm{x}_\sigma,\sigma)$, and we compute the flow matching loss in velocity space,
\begin{equation}
  \mathcal{L}_{\text{flow}} = \mathbb{E}_{\bm{x}, \bm{\epsilon}, \sigma} \big\| \hat{\bm{v}} - \bm{v} \big\|_2^2,
\end{equation}
where $\hat{\bm{v}} = (\bm{x}_\sigma - \hat{\bm{x}}) / \max(\sigma, \sigma_{\min})$ and $\bm{v} = (\bm{x}_\sigma - \bm{x}) / \max(\sigma, \sigma_{\min})$ are the predicted and target velocities, and $\sigma_{\min}$ clips the denominator to avoid numerical instability at low noise levels. 
Latent models typically predict $\bm{v}$ while pixel-space image models either predict $\bm{v}$~\cite{chen2026dip, chen2026l2p, yu2026pixeldit} or switch to $\bm{x}$~\cite{li2026back, jiang2026empirical}.
We find $\bm{x}$-prediction essential for video, as it enables the detailer to render the DiT tokens into clean local structure that lies near a low-dimensional manifold~\cite{li2026back}.
Predicting $\bm{v}$ would instead require the network to model the full-rank per-pixel noise in $\bm{x}_\sigma$, a burden that increases with patch size.
Figure~\ref{fig:abl_pred} shows $\bm{v}$-prediction fills the video with high-frequency, flickering noise.

\vspace{2.5pt}\noindent\textbf{Auxiliary Losses.}
The temporal convolutions of the detailer are trained from scratch, so we add a loss on frame differences to learn temporal smoothness,
\begin{equation}
  \mathcal{L}_{\text{temp}} = \mathbb{E}_{\bm{x}, \bm{\epsilon}, \sigma}\, \rho\big(\Delta\hat{\bm{x}} - \Delta\bm{x}\big),
\end{equation}
where $\Delta\bm{x}^{(f)}=\bm{x}^{(f+1)}-\bm{x}^{(f)}$ is the difference between consecutive frames $f$ and $f+1$, and $\rho(u)=\sqrt{u^2+\delta^2}$ is the Charbonnier penalty with a small constant $\delta$, averaged over all elements.
We also add an LPIPS perceptual loss~\cite{zhang2018unreasonable} on $\hat{\bm{x}}$, which improves the visual quality of pixel-space image models~\cite{ma2026pixelgen}.
Both losses are weighted toward low noise levels, where $\hat{\bm{x}}$ is informative. The total loss is thus
\begin{equation}
  \mathcal{L} = \mathcal{L}_{\text{flow}} + \lambda_{\text{temp}}\,\mathcal{L}_{\text{temp}} + \lambda_{\text{LPIPS}}\,\mathcal{L}_{\text{LPIPS}},
\end{equation}
where $\lambda_{\text{temp}}$ and $\lambda_{\text{LPIPS}}$ are loss weights.

\subsection{Efficient High-Resolution Generation}\label{sec:high_resolution}

\begin{figure}[t]
  \centering
  \includegraphics[width=\columnwidth]{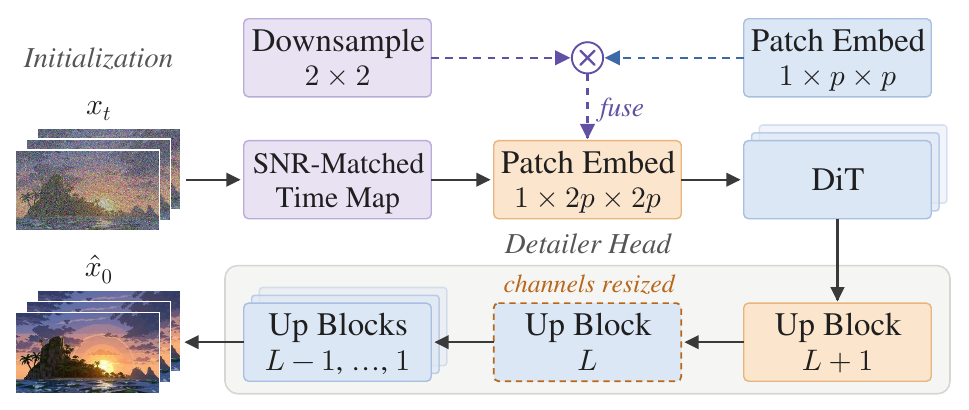}
  \caption{\textbf{Patch-size adaptation.} To obtain $1{\times}2p{\times}2p$ patches, the new patch embedding weight is obtained by fusing $2{\times}2$ average downsampling with the pretrained
  $1{\times}p{\times}p$ patch embedding. A new up block $L{+}1$ is prepended to the detailer head and initialized as nearest neighbor upsampling. With token count unchanged, we train at $2{\times}$ resolution
  at the same DiT cost.}
  \label{fig:patch_adapt}
\end{figure}

The high-level semantic content of a video changes little with resolution, so higher resolution should mainly require finer rendering, not more tokens.
A latent DiT cannot exploit this, as doubling the resolution yields $4\times$ more latent tokens.
\methodname\ instead doubles the patch size from $p$ to $2p$ and adds one up block to the detailer, keeping the token grid unchanged (Fig.~\ref{fig:patch_adapt}).
Unlike PixelDiT~\cite{yu2026pixeldit}, whose token count grows with resolution, or DiP~\cite{chen2026dip} and L2P~\cite{chen2026l2p}, which add no detailer levels, this gives the doubled resolution its own upsampling level without much additional cost.
 
We finetune the patch-$2p$ model from the patch-$p$ model. Its patch embedding fuses $2\times2$ average pooling with the patch-$p$ embedding, so the DiT initially sees a downsampled version of its input, and the new upsampling block starts as an identity pass-through, acting as nearest-neighbor upsampling.
We also double both the calibration scale and the noise shift~\cite{esser2024scaling}, so the DiT sees the same noise schedule as at patch $p$. See App.~\ref{sec:detailer_highres} for details.

\definecolor{vbbestbg}{RGB}{168,218,181}
\definecolor{vbsecondbg}{RGB}{226,242,230}
\newcommand{\vbbest}[1]{\cellcolor{vbbestbg}\textbf{#1}}
\newcommand{\vbsecond}[1]{\cellcolor{vbsecondbg}\underline{#1}}
\newcommand{\vbhead}[1]{{\footnotesize\begin{tabular}[b]{@{}c@{}}#1\end{tabular}}}

\begin{table*}[t]
\centering
\renewcommand{\arraystretch}{1.1}
\setlength{\tabcolsep}{2.2pt}
\resizebox{\textwidth}{!}{%
\begin{tabular}{@{}l c ccc ccccccc ccccccccc@{}}
\toprule
& & \multicolumn{3}{c}{Overall}
  & \multicolumn{7}{c}{Quality dimensions}
  & \multicolumn{9}{c}{Semantic dimensions} \\
\cmidrule(lr){3-5} \cmidrule(lr){6-12} \cmidrule(l){13-21}
Model & Size
  & \vbhead{Total} & \vbhead{Quality} & \vbhead{Semantic}
  & \vbhead{Subject\\consist.} & \vbhead{Backgr.\\consist.} & \vbhead{Temporal\\flicker.} & \vbhead{Motion\\smooth.} & \vbhead{Dynamic\\degree} & \vbhead{Aesthetic\\quality} & \vbhead{Imaging\\quality}
  & \vbhead{Object\\class} & \vbhead{Multiple\\objects} & \vbhead{Human\\action} & \vbhead{Color} & \vbhead{Spatial\\relation.} & \vbhead{Scene} & \vbhead{Appear.\\style} & \vbhead{Temporal\\style} & \vbhead{Overall\\consist.} \\
\midrule
Wan2.1-T2V~\cite{wan2025wan} & 14B
  & 84.05 & 84.91 & 80.63
  & \vbsecond{97.22} & \vbsecond{98.20} & 99.21 & 98.28 & 60.83 & 66.60 & 67.77
  & 95.24 & 86.65 & \vbbest{97.00} & \vbsecond{84.73} & 78.42 & 52.18 & \vbsecond{22.52} & \vbsecond{24.57} & 26.89 \\
PixelUMM~\cite{wei2026pixelumm} & 8B MoT
  & 83.72 & 84.19 & \vbsecond{81.86}
  & \vbbest{97.27} & 96.75 & \vbsecond{99.25} & \vbbest{98.69} & 58.33 & 62.56 & 68.72
  & \vbsecond{96.79} & \vbbest{89.31} & 96.20 & \vbbest{90.24} & 77.38 & 55.35 & 21.86 & \vbbest{24.84} & \vbbest{27.23} \\
\midrule
\methodname\ 480p & 14B
  & \vbsecond{84.78} & \vbsecond{85.44} & \vbbest{82.15}
  & 96.65 & 97.30 & 98.94 & 96.92 & \vbbest{73.89} & \vbsecond{67.30} & \vbsecond{71.25}
  & 95.66 & \vbsecond{88.87} & 96.20 & 81.59 & \vbbest{84.74} & \vbbest{57.75} & \vbbest{23.23} & 24.32 & \vbsecond{26.91} \\
\methodname\ 960p & 14B
  & \vbbest{85.09} & \vbbest{86.19} & 80.68
  & 97.00 & \vbbest{98.51} & \vbbest{99.59} & \vbsecond{98.31} & \vbsecond{65.00} & \vbbest{67.49} & \vbbest{71.83}
  & \vbbest{97.59} & 88.16 & \vbsecond{96.40} & 74.54 & \vbsecond{84.14} & \vbsecond{56.22} & 21.91 & 24.46 & 26.58 \\
\bottomrule
\end{tabular}%
}
\caption{\textbf{VBench comparison.}
\methodname\ is finetuned from the latent model Wan2.1-T2V-14B; PixelUMM is a concurrent pixel-space model.
Each model generates five videos per GPT-enhanced prompt, and we score the first $1.3$\,s, the length of our videos (App.~\ref{sec:app_t2v}).
\methodname\ has a higher total score than both baselines at both resolutions, so lifting generation from latents to pixels does not degrade quality; its largest gains over Wan2.1 are in dynamic degree, spatial relationship, scene, and imaging quality, and its largest loss is in color.
{\setlength{\fboxsep}{1pt}\colorbox{vbbestbg}{\textbf{Bold}}: best; \colorbox{vbsecondbg}{\underline{underlined}}: second best.}}
\label{tab:vbench}
\end{table*}

\subsection{Precise Video Editing}\label{sec:video_editing}

Precise video editing, such as re-rendering a video from its per-frame depth, requires the generated video to align with the condition at every frame.
Latent models typically encode the condition with the same VAE as the video~\cite{vace}, so several condition frames share one latent, leading to inaccurate alignment, especially in fast motions.
In \methodname, the condition is tokenized per frame as the corresponding video, so generation and condition are aligned at the frame level, without changing the conditioning mechanism.
Concretely, we add the control branch of VACE~\cite{vace}: conditions are residually added to the DiT tokens via control blocks, and the condition's patch embedding is initialized, like the video's, by fusing the pretrained embedding with $\bm{A}$.

\subsection{Low-Latency Autoregressive Generation}\label{sec:autoregressive_gen}

\begin{figure*}[!t]
  \centering
  \includegraphics[width=\textwidth]{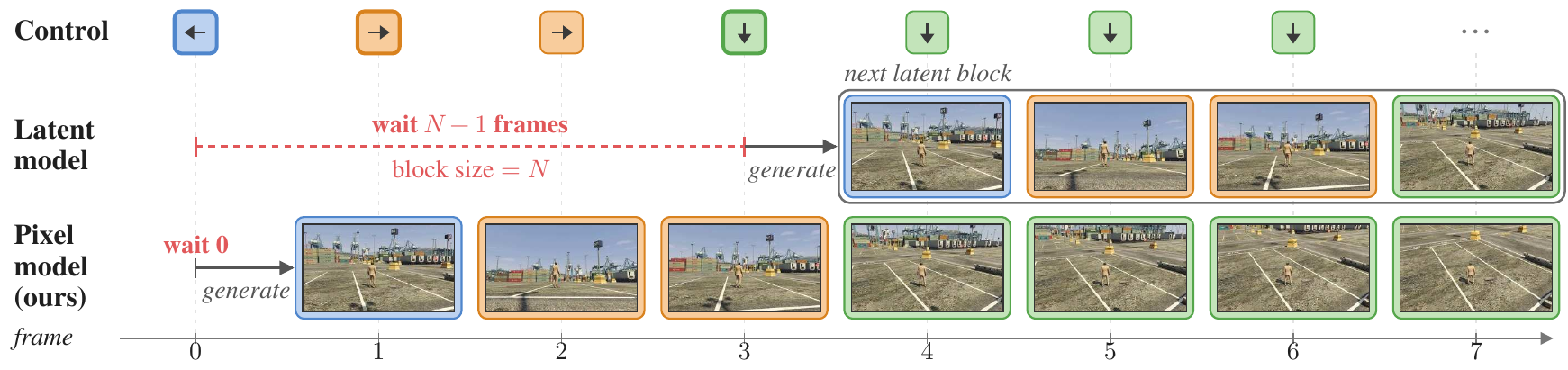}
  \caption{\textbf{Control-to-screen latency.}
    Control inputs, \eg turn right ($\rightarrow$), turn left ($\leftarrow$), and look down ($\downarrow$), arrive over time (top row). Colors indicate which control each generated frame reflects.
    \emph{Latent model:} controls are reflected in the next block. With block size $N$ this means a wait of up to $N-1$ frames. We use $N=4$ for this visualization.
    \emph{Pixel model (ours):} generates one frame per control, so controls are reflected with zero wait, making it better suited for real-time interactive generation.
    }
  \label{fig:latency}
\end{figure*}

Interactive applications stream video in response to user controls, such as camera or motion inputs, and should react to each control as soon as it arrives.
An autoregressive model generates video in blocks, and a block can start only once the controls for all its frames have arrived.
In a temporally compressed latent model, each block spans at least one latent, \ie, several frames, so a new control waits for the rest of its block, a delay that no throughput-side optimization can remove (Fig.~\ref{fig:latency}).
\methodname\ can generate single-frame blocks, so each control takes effect at the next frame.
To achieve this, we lift a bidirectional latent model to pixel space with the recipe above and distill it into a few-step causal model with Self Forcing~\cite{huang2026self}.
The causal detailer decodes one frame at a time from cached features of previous frames, keeping its cost per frame constant.
Single-frame blocks trade some throughput for this latency gain, which we analyze in Sec.~\ref{sec:ar-exp}.

\section{Experiments}

We evaluate \methodname\ on text-to-video generation in Sec.~\ref{sec:t2v_exp}, followed by its three applications: efficient high-resolution generation in Sec.~\ref{sec:highres_exp}, precise video editing in Sec.~\ref{sec:editing_exp}, and low-latency autoregressive generation in Sec.~\ref{sec:ar-exp}.
Finally, we ablate our key design choices in Sec.~\ref{sec:ablation}.

\subsection{Experimental Setup}
\label{sec:setup}
 
We lift three pretrained latent models into pixel space: Wan2.1-T2V-14B~\cite{wan2025wan} for text-to-video and high-resolution generation, Wan2.1-T2V-1.3B for video editing and ablations, and the camera-controlled SolarWM-5B~\cite{huang2026solarwmopendatascalable} for interactive generation.
Unless otherwise noted, models generate $21
$-frame videos with patch size $p=16$ and sample with $50$ Euler steps at the training shift.
Each experiment below gives its key setup. Full details are in App.~\ref{sec:app_exp}.

\subsection{Text-to-Video Generation}
\label{sec:t2v_exp}

\noindent\textbf{Setup.}
We finetune Wan2.1-T2V-14B~\cite{wan2025wan} into a frame-wise pixel-space model at $480\times832$ for $20$K iterations, on $64$K videos generated by Wan2.1.
We use shift $s=11$, which gives the DiT nearly the same effective shift as the best 1.3B setting (Sec.~\ref{sec:ablation}).
We evaluate on VBench~\cite{huang2024vbench} with its $946$ GPT-enhanced prompts and five videos per prompt, and run a pairwise human study on $100$ random prompts from Movie Gen Video Bench~\cite{polyak2024movie} comparing visual quality, motion quality, and text alignment.
We compare against our base latent model, Wan2.1-T2V-14B, and the concurrent pixel-space model PixelUMM~\cite{wei2026pixelumm} using their official settings (App.~\ref{sec:app_t2v}).
Since our models generate $21$ frames, we score every model on the first $1.3$\,s of each video, which has no meaningful impact compared to scoring full videos in \cref{tab:vbench_full}.

\begin{figure}[t]
  \centering
  \includegraphics[width=\columnwidth]{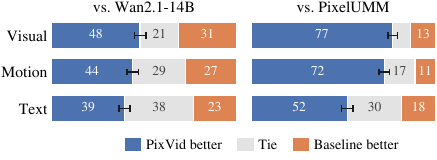}
  \caption{\textbf{Human evaluation.} Pairwise preferences between \methodname\ (480p) and baselines on $100$ Movie Gen~\cite{polyak2024movie} prompts, judged by $10$ evaluators. Error bars indicate $95\%$ confidence intervals on \methodname's share of votes. \methodname\ wins in all comparisons.}
  \label{fig:ab}
\end{figure}

\vspace{2.5pt}\noindent\textbf{Results.}
In \cref{tab:vbench}, \methodname\ reaches a VBench total of $84.78$, above both Wan2.1 ($84.05$) and PixelUMM ($83.72$).
Replacing the VAE decoder with the detailer head does not cost visual quality: imaging quality, aesthetic quality, and dynamic degree improve over Wan2.1, while consistency and smoothness stay within $1.4$ points.
\methodname\ also achieves the highest semantic score of $82.15$; color is the only semantic dimension where it trails Wan2.1 by more than a point ($81.59$ vs.\ $84.73$).
In the human study in \cref{fig:ab}, raters prefer \methodname\ across all comparisons, showing that finetuning is competitive with the pretrained latent model and compares favorably with PixelUMM, which is trained from scratch.

\subsection{Efficient High-Resolution Generation}
\label{sec:highres_exp}

\noindent\textbf{Setup.}
We generate $960\times1664$ videos with the same $30\times52$ token grid per frame as $480\times832$ generation by doubling the patch size from $16$ to $32$.
We first train a patch-16 pixel model at $480\times832$, then finetune it at $960\times1664$ with patch $32$ with twice the shift and an additional scale-$2$ calibration.
We train on $40$K synthetic videos resized and cropped to the training resolution, and evaluate with the same VBench protocol as \cref{sec:t2v_exp}.
 
\vspace{2.5pt}\noindent\textbf{Results.}
Using the same number of tokens per frame as $480$p, the $960\times1664$ model reaches the highest VBench total of $85.09$ and the best imaging, aesthetic, background-consistency, and temporal-flickering scores in \cref{tab:vbench}. Its semantic score of $80.68$ is below the $480\times832$ model's, mostly in color.
Doubling the patch size therefore keeps \methodname\ competitive at $4$ times the pixels per frame.
Its cost per clip stays close to a latent model's at this resolution: a $21$-frame clip has $32,760$ tokens and takes $192$\,s on one B200, against $37,440$ tokens and $176$\,s for a Wan2.1-VAE latent model (App.~\ref{sec:app_cost}).
Since VBench resizes frames to at most $512$ pixels before scoring, it does not measure the added detail, which we show in Fig.~\ref{fig:teaser}(a).

\subsection{Precise Video Editing}
\label{sec:editing_exp}

\begin{table}[t]
  \centering
  \footnotesize
  \renewcommand{\arraystretch}{1.12}
  \setlength{\tabcolsep}{1.7pt}
    \begin{tabular}{lcccccc}
    \toprule
    & \multicolumn{3}{c}{Alignment} & \multicolumn{2}{c}{Temporal} & Quality \\
    \cmidrule(lr){2-4}\cmidrule(lr){5-6}\cmidrule(lr){7-7}
    Model & Motion$\downarrow$ & Kept$\uparrow$ & Depth$\downarrow$ & tLP$\downarrow$ & FVMD$\downarrow$ & FVD$\downarrow$ \\
    \midrule
    \multicolumn{7}{l}{\emph{Zero-shot}} \\
    VACE-1.3B~\cite{vace} & 30.4 & 32\% & 0.270 & 0.075 & 3119 & 199 \\
    VACE-14B~\cite{vace} & 24.7 & 63\% & 0.268 & 0.055 & 2469 & 250 \\
    Wan2.1-Fun-1.3B~\cite{aigc_apps_VideoX_Fun_2026} & 27.3 & 56\% & 0.275 & 0.057 & 2453 & 343 \\
    Wan2.2-Fun-5B~\cite{aigc_apps_VideoX_Fun_2026} & 32.4 & 46\% & 0.344 & 0.057 & 2921 & 239 \\
    Cosmos-2.5-2B~\cite{nvidia2025worldsimulationvideofoundation} & 24.6 & 51\% & 0.221 & 0.048 & 2349 & 609 \\
    \midrule
    \multicolumn{7}{l}{\emph{Finetuned on our data}} \\
    Latent (VACE-1.3B) & 17.9 & 83\% & 0.136 & 0.039 & 1678 & \textbf{91} \\
    \methodname & \textbf{10.7} & \textbf{98\%} & \textbf{0.109} & \textbf{0.030} & \textbf{1512} & 118 \\
    \bottomrule
  \end{tabular}
  \caption{\textbf{Precise video editing.} Motion error in pixels, motion kept, depth error, and temporal LPIPS (tLP) are medians over the $250$ fast clips, with all motion bins in \cref{tab:app_d2v}; FVMD and FVD are computed over all clips (App.~\ref{sec:app_d2v}).}
  \label{tab:d2v}
\end{table}

\noindent\textbf{Setup.}
We add the control branch of VACE~\cite{vace} to \methodname-1.3B and finetune it to re-render a video from its per-frame depth and a caption.
We compare with VACE-1.3B, finetuned with the same control branch, data, and training settings;
and with off-the-shelf depth-conditioned models without finetuning.
We train on $79$K stock videos with varied camera and object motion and paired depth~\cite{chen2025video}, and evaluate on $1{,}000$ held-out clips across alignment, temporal fidelity, and video quality.
Alignment is measured by optical flow error, the fraction of source motion kept, and depth error, while temporal fidelity and video quality are measured by temporal LPIPS, FVMD~\cite{liu2024fvmd}, and FVD~\cite{unterthiner2018towards}.
Fast clips are the quarter with the most motion; details are in App.~\ref{sec:app_d2v}.

\begin{figure}[t]
  \centering
  \includegraphics[width=\columnwidth]{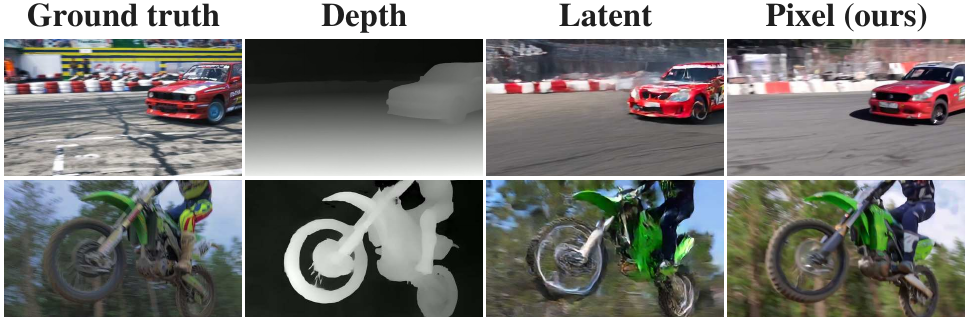}
  \caption{\textbf{Depth-conditioned Editing.} Our pixel model enables precise video editing under high motion, while the temporally compressed latent model morphs and blurs fast-moving objects.}
  \label{fig:d2v}
\end{figure}

\vspace{2.5pt}\noindent\textbf{Results.}
\methodname\ keeps the scene intact under fast motion, where the latent model morphs and blurs moving objects (Fig.~\ref{fig:d2v}).
On fast clips, it lowers motion error by $40\%$ and depth error by $20\%$, and keeps $98\%$ of the source motion, against $83\%$ for the finetuned latent model and at most $63\%$ for the zero-shot models (\cref{tab:d2v}).
It also has the lowest FVMD and tLP and leads on all four per-clip metrics over all $1{,}000$ clips (\cref{tab:app_d2v}), while its FVD trails ($118$ vs.\ $91$).

\subsection{Low-Latency Autoregressive Generation}
\label{sec:ar-exp}

\noindent\textbf{Setup.}
We lift the bidirectional SolarWM-5B base~\cite{huang2026solarwmopendatascalable} into pixel space and distill it with Self Forcing~\cite{huang2026self} into a $4$-step causal student that generates one frame per block.
We compare with the official causal student distilled from the same base, referred to as SolarWM, which generates blocks of $12$ frames ($3$ latents). 
We evaluate quality and camera following on camera paths from nine public datasets, and responsiveness on abrupt turns, measuring how closely the generated motion follows each control (App.~\ref{sec:app_ar}).
We measure control-to-screen latency, which includes both the wait for controls from the same block and the model's generation time. We time both models on one B200 GPU with the same kernels.

\vspace{2.5pt}\noindent\textbf{Results.}
\methodname\ cuts median control-to-screen latency from $495$\,ms to $34$\,ms (\cref{tab:ar}).
Most of SolarWM's latency is waiting for controls from the same $12$-frame block, so at $16$\,fps a control waits up to $688$\,ms before generation even begins.
A one-latent block would still wait $3$ frames, and our experiments find finetuning SolarWM to one-latent blocks hurts performance (App.~\ref{sec:app_1latent}).
Frame-wise generation lowers throughput, since every frame has its own tokens: \methodname\ needs about $3\times$ SolarWM's DiT FLOPs per frame and runs at $29.4$ vs.\ $72.0$\,fps, but both models exceed their base Wan model's native $16$\,fps, so both stream in real time.
SolarWM's extra throughput cannot shorten its control-to-screen latency.

\begin{table}[t]
  \centering
  \footnotesize
  \renewcommand{\arraystretch}{1.12}
  \setlength{\tabcolsep}{1.5pt}
  \begin{tabular}{lcccccccc}
    \toprule
    & \multicolumn{4}{c}{Latency and speed} & Quality & \multicolumn{3}{c}{Camera} \\
    \cmidrule(lr){2-5}\cmidrule(lr){6-6}\cmidrule(lr){7-9}
    Model & Block & Wait$\downarrow$ & Meas.$\downarrow$ & FPS$\uparrow$ & FVD$\downarrow$ & Rot.$\downarrow$ & Tr.$\downarrow$ & Turn$\uparrow$ \\
    \midrule
    SolarWM~\cite{huang2026solarwmopendatascalable} & 12 & 11 & 495\,ms & \textbf{72.0} & 53.8 & \textbf{4.24$^\circ$} & \textbf{0.231} & 0.70 \\
    \methodname & \textbf{1} & \textbf{0} & \textbf{34\,ms} & 29.4 & \textbf{51.1} & 5.44$^\circ$ & 0.312 & \textbf{0.95} \\
    \bottomrule
  \end{tabular}
  \caption{\textbf{Low-latency autoregressive generation.}
    Wait and Meas.\ report the maximum control wait in frames and median control-to-screen latency on one B200; Rot., Tr., and Turn measure camera rotation error, translation error, and motion correlation on abrupt turns (App.~\ref{sec:app_ar}).
    Both models exceed their native $16$\,fps, but only \methodname\ responds at the next frame, with $14\times$ lower control-to-screen latency. }
  \label{tab:ar}
\end{table}

\begin{figure}[t]
  \centering
  \includegraphics[width=\columnwidth]{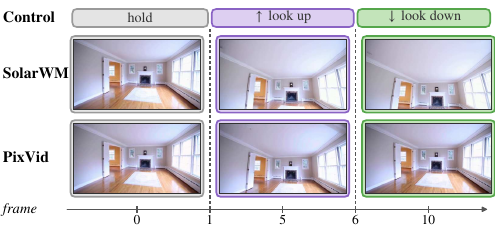}
  \caption{\textbf{Responsiveness.} SolarWM receives one pose per four-frame latent, resulting in inaccurate motion when the direction changes; here continuing to look up when it should look down. \methodname\ handles direction changes natively.}
  \label{fig:ar_turn}
\end{figure}

\methodname\ follows abrupt turns more closely (Turn $0.95$ vs.\ $0.70$; \cref{fig:ar_turn}), since it receives a camera pose for every frame, whereas SolarWM receives one per four-frame latent.
Its FVD is comparable to SolarWM's ($51.1$ vs.\ $53.8$), while its camera rotation and translation errors are higher ($5.44^\circ$ vs.\ $4.24^\circ$ and $0.312$ vs.\ $0.231$). 
This is likely due to drift accumulated over a longer sequence of poses. Indeed, at $21$ frames our errors are comparable to SolarWM’s (\cref{tab:app_horizon}). Mitigating such drift is an active area of research, so we leave it to future work.

\subsection{Ablation Study}
\label{sec:ablation}

\noindent\textbf{Setup.}
All ablations finetune Wan2.1-T2V-1.3B for $10$K iterations and change one design choice at a time from our final recipe: the convolutional head, $\bm{x}$-prediction, and shift.
We generate one video for each of $119$ VBench prompts and measure the artifact that each choice targets using three reference-free metrics defined in App.~\ref{sec:app_abl}: Seam, discontinuities along the patch grid; Noise, the per-pixel noise level; and Flicker, noise that changes from frame to frame.
The noise shift decides how training and sampling are spread across noise levels, which affects global structure as well as local noise, so we score the shift sweep on VBench, with all $946$ prompts and one video per prompt.

\begin{figure}[t]
\centering
\begingroup
\setlength{\tabcolsep}{2pt}
\newcommand{\ablimg}[1]{\includegraphics[width=0.488\linewidth]{figures/images/ablation-sampling/#1}}
\begin{tabular}{@{}cc@{}}
{\scriptsize\textbf{Linear head}} & {\scriptsize\textbf{Convolutional head (ours)}} \\[-1pt]
{\scriptsize {\color{gray}median patch seams:} \textcolor{red}{\textbf{118.6}}} & {\scriptsize {\color{gray}median patch seams:} \textbf{13.3}} \\[3pt]
\ablimg{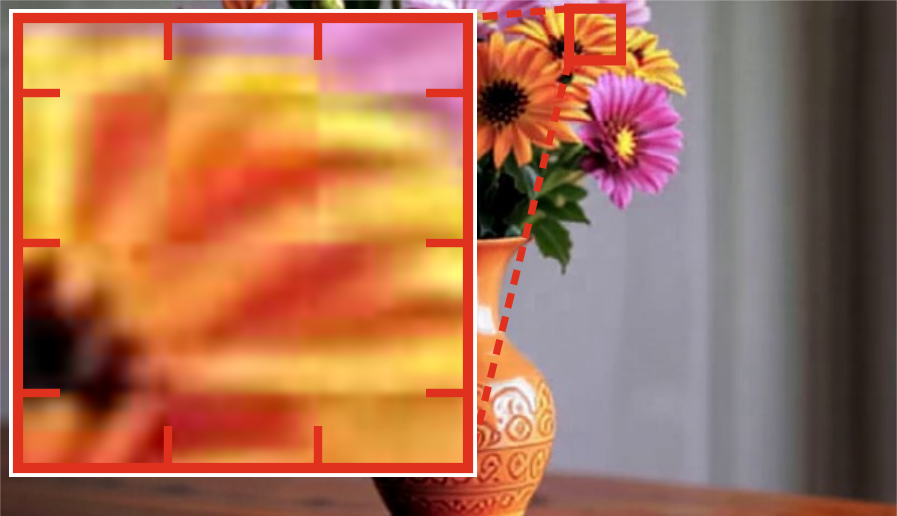} & \ablimg{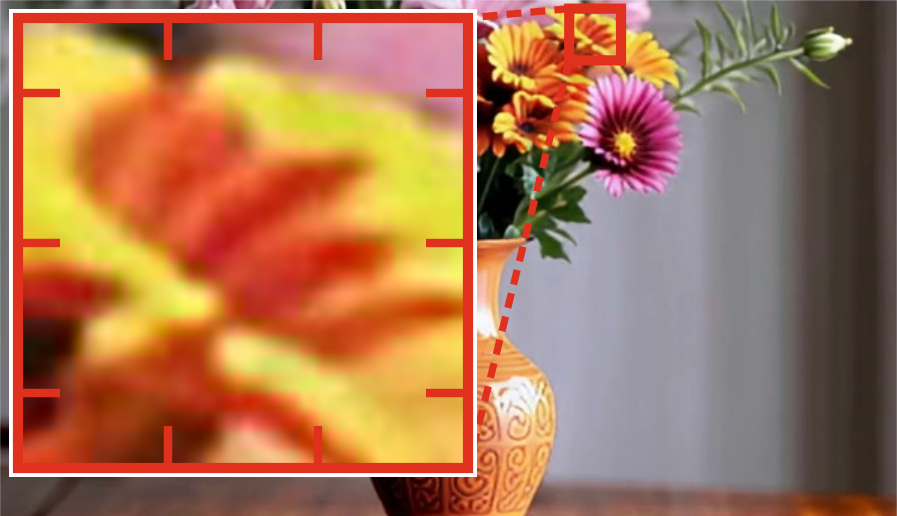} \\[3pt]
\ablimg{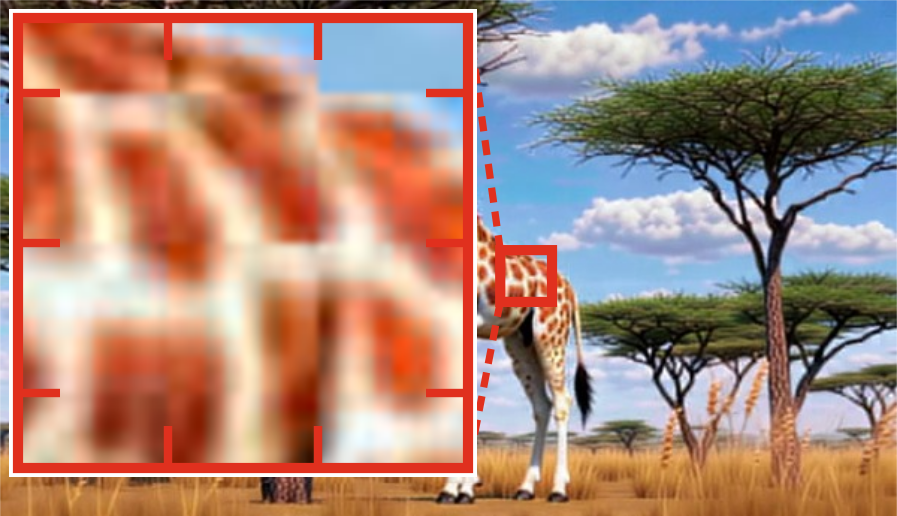} & \ablimg{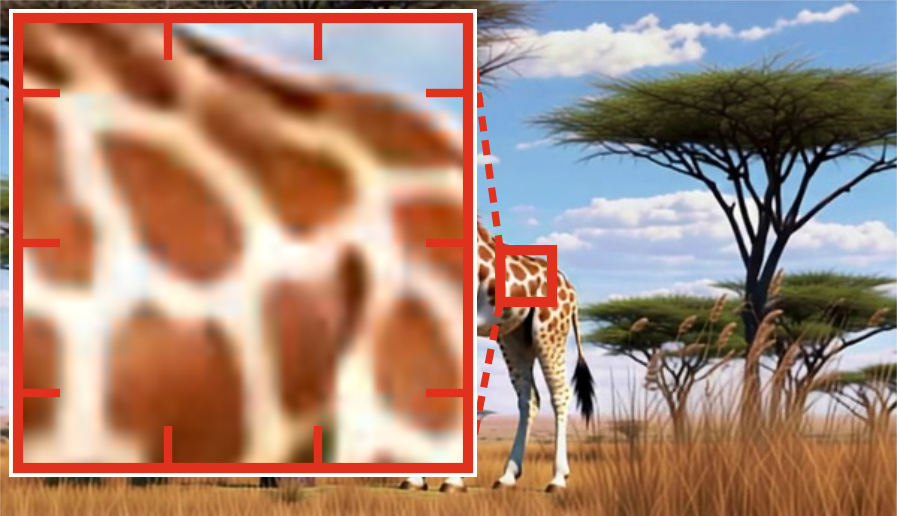} \\
\end{tabular}
\endgroup
\caption{\textbf{Impact of output head.} Under $\bm{x}$-prediction, a linear head maps each DiT token to its pixel patch independently, causing visible grid-like artifacts, while our convolutional head removes most seams. Metric details are in App.~\ref{sec:app_abl}.}
\label{fig:abl_head}
\end{figure}

\vspace{2.5pt}\noindent\textbf{Results.}
\noindent\emph{Output head.} \label{sec:exp_ablation_head}
A linear head maps each DiT token to its pixel patch independently, so the head itself does not make neighboring patches agree at their borders, and the patch grid stays visible 
(Fig.~\ref{fig:abl_head}).
Our convolutional head, whose kernels span patch borders, lowers the seam measure from $118.6$ to $13.3$.
The U-Net head, which additionally encodes the noisy input, reaches a similar seam ($9.2$) with slightly higher noise ($0.848$ vs.\ $0.777$) and flicker ($0.429$ vs.\ $0.406$), and it has $5.6\times$ more parameters ($248$M vs.\ $44$M at $1.3$B), so we use the decoder-only head.

\begin{figure}[t]
\centering
\begingroup
\setlength{\tabcolsep}{2pt}
\newcommand{\ablimg}[1]{\includegraphics[width=0.488\linewidth]{figures/images/ablation-sampling/#1}}
\begin{tabular}{@{}cc@{}}
{\scriptsize\textbf{$\bm{v}$-prediction}} & {\scriptsize\textbf{$\bm{x}$-prediction (ours)}} \\[-1pt]
{\scriptsize {\color{gray}median pixel noise:} \textcolor{red}{\textbf{62.25}}} & {\scriptsize {\color{gray}median pixel noise:} \textbf{0.78}} \\[3pt]
\ablimg{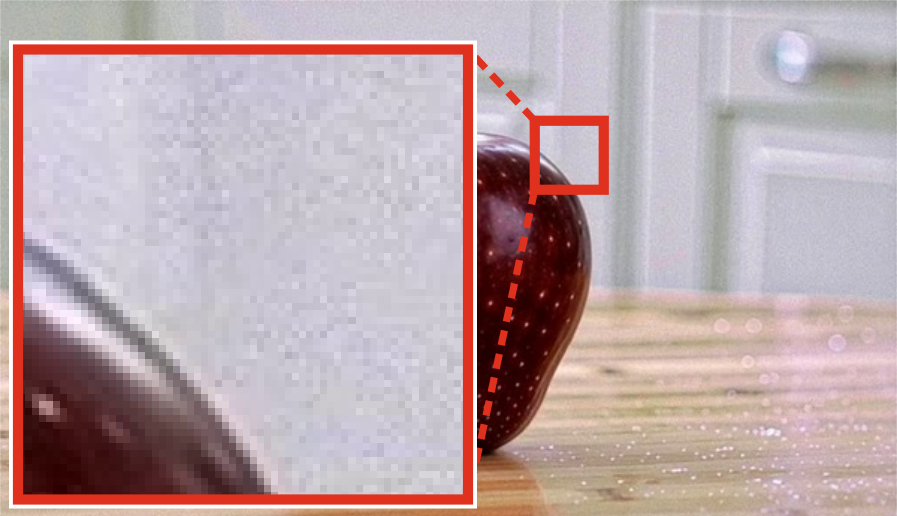} & \ablimg{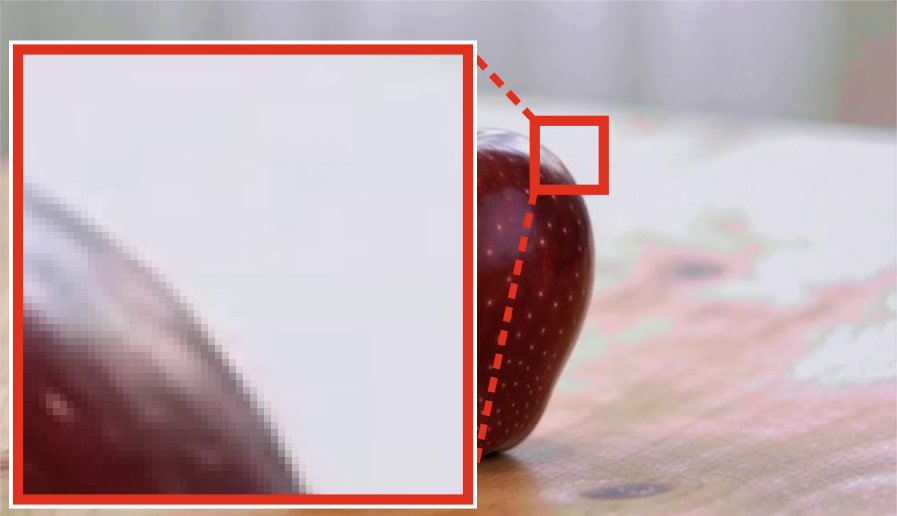} \\[3pt]
\ablimg{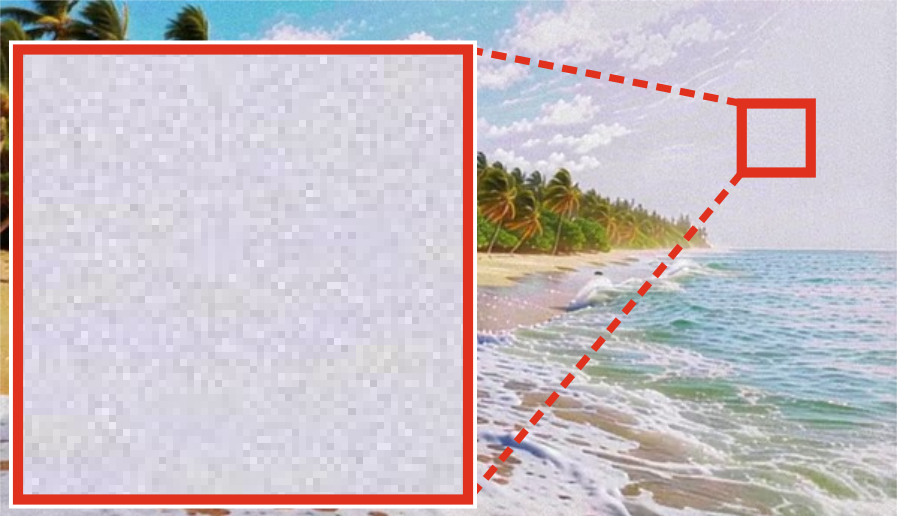} & \ablimg{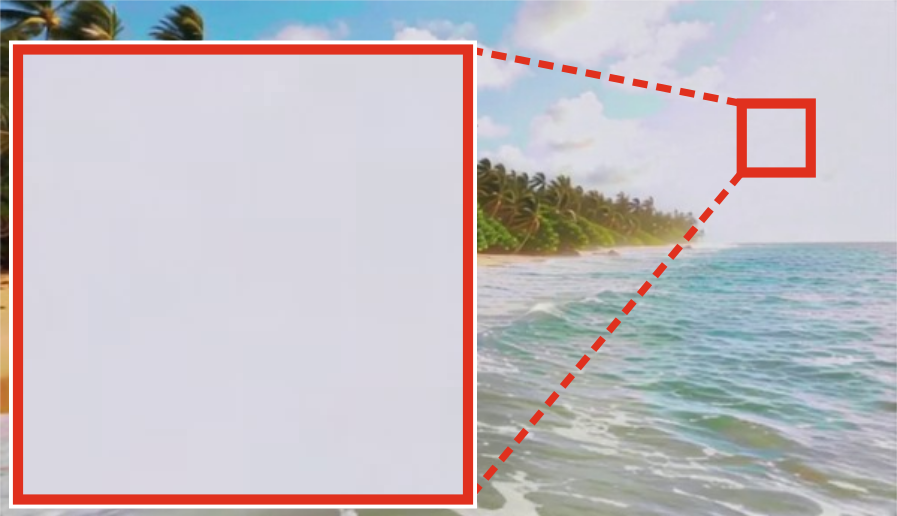} \\
\end{tabular}
\endgroup
  \caption{\textbf{Impact of prediction target.} With the same convolutional head, $\bm{v}$-prediction produces grainy frames at $80\times$ the estimated noise of $\bm{x}$-prediction, while $\bm{x}$-prediction produces clean ones. Metric details are in App.~\ref{sec:app_abl}.}
\label{fig:abl_pred}
\end{figure}

\vspace{2.5pt}\noindent\emph{Prediction target.}
As shown in \cref{fig:abl_pred}, $\bm{v}$-prediction with our head fills the video with flickering high-frequency noise at $80$ times the noise level of $\bm{x}$-prediction, because predicting velocity requires the network to reproduce the per-pixel noise of its input.
Asymmetric $\bm{v}$-prediction~\cite{chen2026asymmetric} reduces
the seams of the linear head ($24.5$ vs.\ $118.6$), but its linear recovery rule still leaves nearly twice the seams of our head, and the noisy input it reads raises noise and flicker ($0.897$ and $0.524$) above those of $\bm{x}$-prediction.

\vspace{2.5pt}\noindent\emph{Noise shift.}
Since pixels are more redundant than latents, the SNR-aligned timestep calibration in App.~\ref{sec:patch_embed_init} naturally dictates a higher shift for pixel models. Here we ablate the uncalibrated shift $s$.

\Cref{tab:abl_shift} shows that VBench peaks at $s=15$ in both \emph{total} and \emph{semantic} score, with $s=5$ close behind in total ($84.27$ vs.\ $84.40$), and declines at larger shifts, mostly in spatial relationship and multiple objects. We therefore use $s=15$, giving an effective shift of $3.97$ under scale $c=3.78$. For the other models we keep a similar effective shift: $s=11$ for the 14B model gives an effective shift of $3.86$ under $c=2.85$, and at $960\times1664$ we double both the shift and the calibration.

\begin{figure}[t]
\centering
\begingroup
\setlength{\tabcolsep}{2pt}
\newcommand{\sampleimg}[1]{\includegraphics[width=0.235\linewidth]{figures/images/ablation-sampling/#1}}
\setlength{\fboxsep}{0pt}
\setlength{\fboxrule}{0.5pt}
\newcommand{\badsampleimg}[1]{\fcolorbox{red}{white}{\includegraphics[width=\dimexpr0.235\linewidth-2\fboxrule\relax]{figures/images/ablation-sampling/#1}}}
\begin{tabular}{@{}cccc@{}}
\multicolumn{2}{c}{\scriptsize\textbf{UniPC} {\color{gray}(median color kept)}} & \multicolumn{2}{c}{\scriptsize\textbf{Euler (ours)} {\color{gray}(median color kept)}} \\[-1pt]
{\scriptsize fp32: 100\%} & {\scriptsize bf16: \textcolor{red}{\textbf{91\%}}} & {\scriptsize fp32: 100\%} & {\scriptsize bf16: \textbf{101\%}} \\[3pt]
\sampleimg{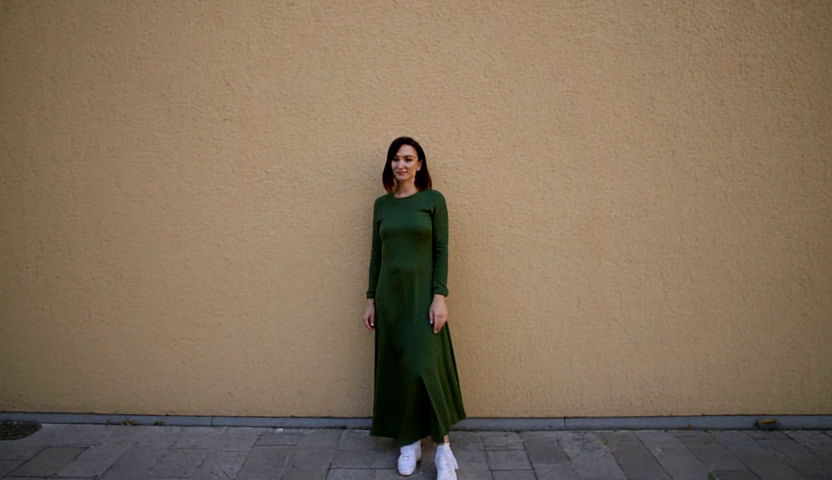} &
\badsampleimg{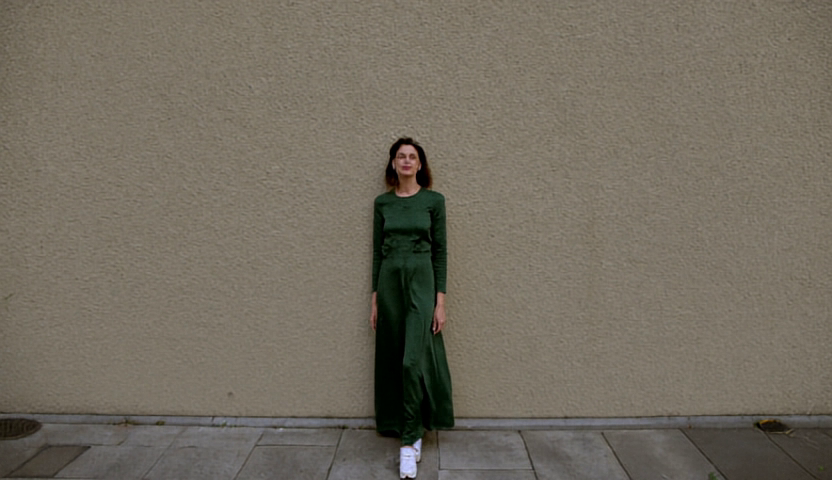} &
\sampleimg{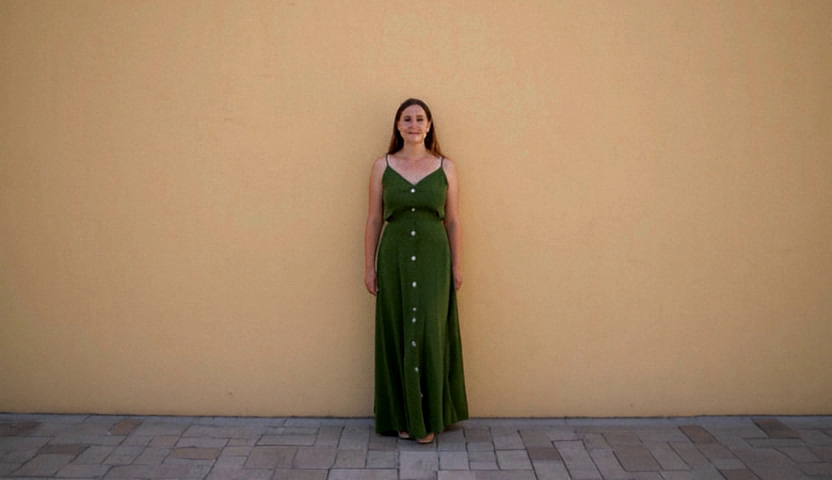} &
\sampleimg{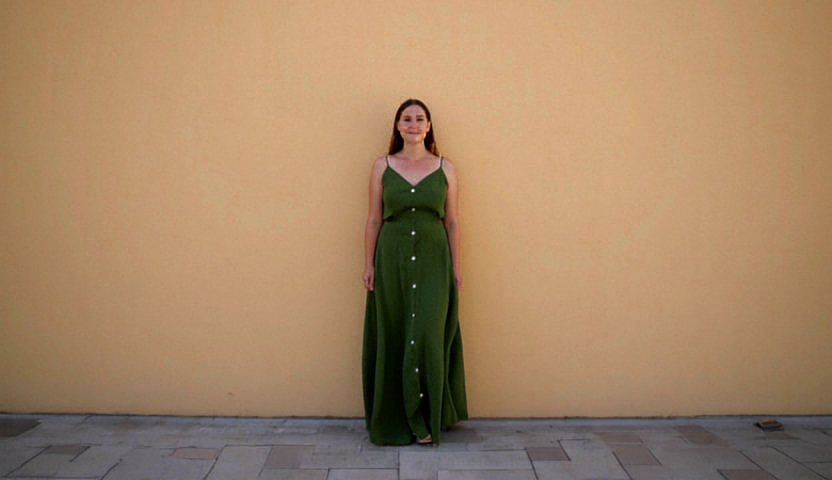} \\[3pt]
\sampleimg{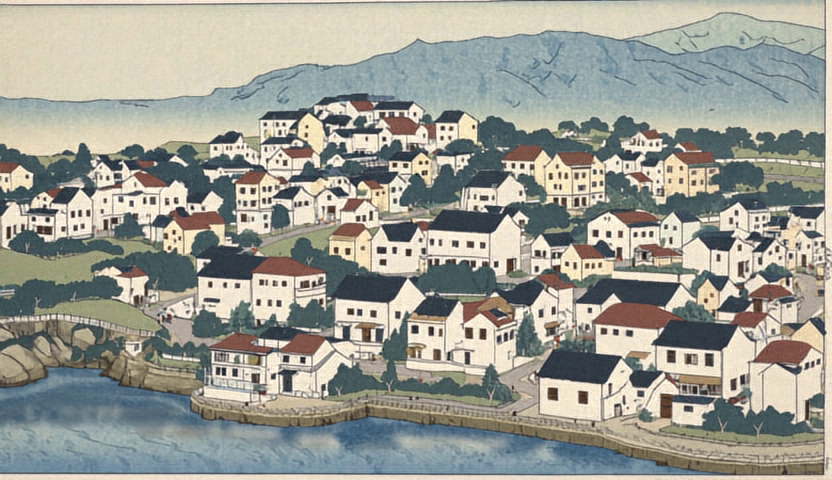} &
\badsampleimg{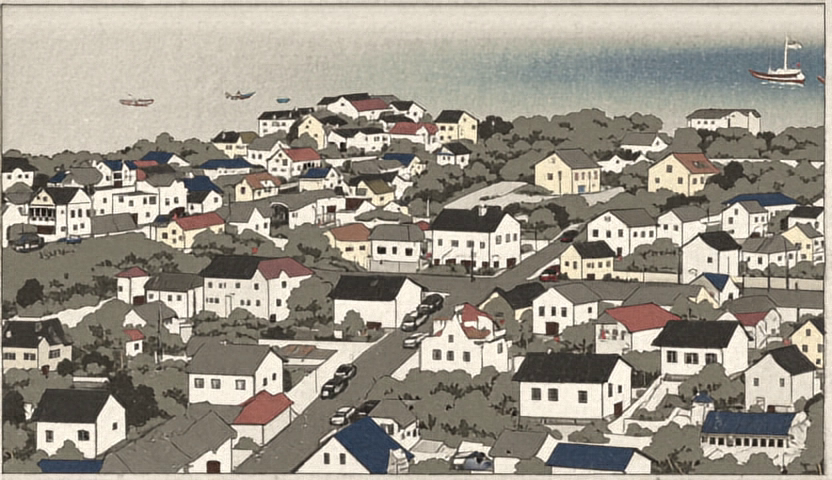} &
\sampleimg{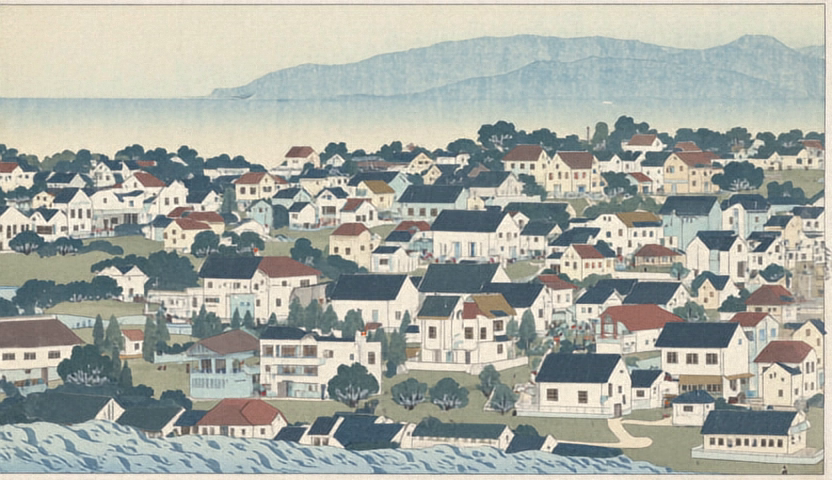} &
\sampleimg{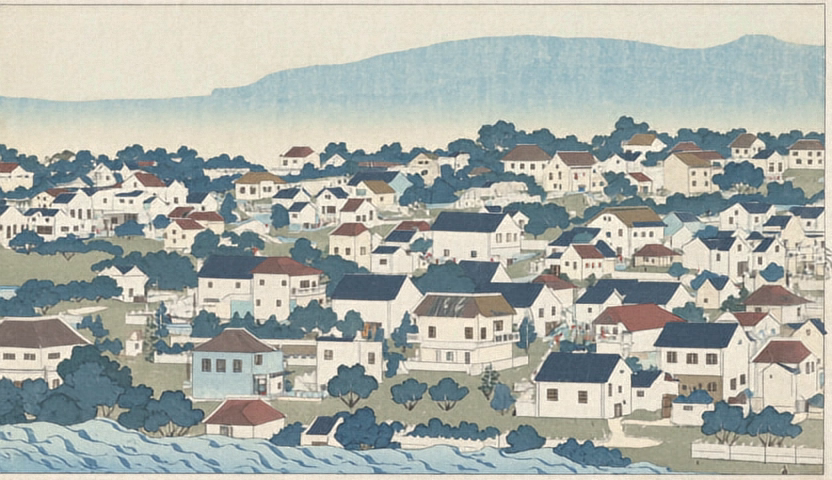} \\
\end{tabular}
\endgroup
\caption{\textbf{Sampler-state precision.}
We use the same prompt and initial noise in every column. With a bf16 sampler state, UniPC drains pixel-space sample color (red), while Euler keeps it.
\emph{Color kept} is mean OKLab chroma relative to an fp32 sampler state.} 
\label{fig:abl_sampler}
\end{figure}

\vspace{2.5pt}\noindent\emph{Sampler.}
Latent video models, including Wan2.1, typically sample with UniPC~\cite{zhao2023unipc}, which works well with a bf16 sampler state in latent space.
In pixel space, a bf16 state makes UniPC drain color, keeping only $91\%$ of the fp32 chroma, whereas Euler keeps it without needing an fp32 state (\cref{fig:abl_sampler}). We suspect pixel space colors are more sensitive to higher order samplers. We therefore use the Euler sampler and leave further analysis to future work.

\begin{table}[t]
  \centering
  \small
  \setlength{\aboverulesep}{0pt}
  \setlength{\belowrulesep}{0pt}
  \renewcommand{\arraystretch}{1.15}
  \begin{tabular*}{\columnwidth}{@{\extracolsep{\fill}}l ccccc@{}}
    \toprule
    Noise shift $s$ & 1 & 5 & \cellcolor{gray!15}15 & 30 & 60 \\
    Effective shift $s/c$ & 0.26 & 1.32 & \cellcolor{gray!15}3.97 & 7.94 & 15.87 \\
    \midrule
    VBench total $\uparrow$ & 81.85 & 84.27 & \cellcolor{gray!15}\textbf{84.40} & 84.09 & 83.42 \\
    VBench quality $\uparrow$ & 83.87 & 85.71 & \cellcolor{gray!15}85.64 & \textbf{85.98} & 85.06 \\
    VBench semantic $\uparrow$ & 73.79 & 78.51 & \cellcolor{gray!15}\textbf{79.46} & 76.49 & 76.88 \\
    \bottomrule
  \end{tabular*}
  \caption{\textbf{Noise shift.}
    Wan2.1-T2V-1.3B with our convolutional head and $\bm{x}$-prediction, trained and sampled with shift $s$; VBench uses all $946$ prompts with one video each.
    The timestep calibration acts as a shift by $1/c$ ($c=3.78$), so the DiT sees the effective shift $s/c$.
    The total and semantic scores peak at $s=15$.}
  \label{tab:abl_shift}
\end{table}

\section{Conclusion}

Latent video diffusion models compress video in space and time with a VAE.
\methodname\ removes this compression by efficiently finetuning a pretrained latent model to generate every frame directly in pixels, reaching quality competitive with the original model.
Working frame by frame in pixels, it overcomes three structural limitations of latent models: resolution scaling, control alignment under fast motion, and control-to-screen latency.

\vspace{3pt}\noindent\textbf{Limitations and future work.}
Without temporal compression, \methodname\ processes $3.9\times$ as many tokens per second of video as Wan2.1 at $480\times832$, so short low-resolution clips cost more than with a latent model, although the gap closes at higher resolution and streaming stays faster than real time (App.~\ref{sec:app_cost}).
Generating directly in pixels also opens up training options that a frozen VAE rules out. Losses and rewards can be defined on pixels, enabling adversarial training and RL with pixel-space rewards. Because all models then share the same output space, distillation across architectures or from mixed teachers also becomes possible.

\vspace{3pt}\noindent\textbf{Acknowledgments.}
We thank Yicong Hong, Zhanpeng Luo, and Wei-An Lin for their assistance in human evaluation.

{
    \small
    \bibliographystyle{ieeenat_fullname}
    \bibliography{main}
}

\clearpage
\appendix

\section{Architecture Details}
\label{sec:architecture_details}
In this section, we show the model architecture details.

\subsection{Input Patch Embedding Initialization}
\label{sec:patch_embed_init}
The pretrained DiT reads latent tokens through a linear patch embedding $\boldsymbol{W}$, while \methodname\ reads pixel patches. Following AsymFlow~\cite{chen2026asymmetric}, we bridge the two with a linear lift between latent tokens and pixel patches, compose it with $\boldsymbol{W}$ to initialize the pixel patch embedding, and calibrate the timestep so that the DiT initially receives inputs on the latent trajectory it was trained on.

\paragraph{Pixel-Latent Alignment.}
Let $\boldsymbol{x} \in \mathbb{R}^{3p^2}$ be a $1\times p\times p$ pixel patch in normalized OKLab, and $\boldsymbol{z} \in \mathbb{R}^{d}$ the latent token at the same location, \ie, the VAE latent patchified by the DiT. For Wan2.1~\cite{wan2025wan}, the $8\times$ spatial VAE downsampling and the $1\times2\times2$ DiT patch give $p=16$, so $3p^2 = 768$ and $d = 64$. We approximate the VAE encoder by a linear lift $\boldsymbol{x} \approx c\,\boldsymbol{A}\boldsymbol{z}$, where $\boldsymbol{A} \in \mathbb{R}^{3p^2\times d}$ has orthonormal columns ($\boldsymbol{A}^\top\boldsymbol{A} = \boldsymbol{I}_d$) and $c > 0$ is a scalar. Collecting $N$ paired patches and tokens from $500$ training clips into $\boldsymbol{X} \in \mathbb{R}^{3p^2\times N}$ and $\boldsymbol{Z} \in \mathbb{R}^{d\times N}$, we fit $\boldsymbol{A}$ by orthogonal Procrustes~\cite{schonemann1966generalized},
\begin{equation}
  \boldsymbol{A} = \argmin_{\boldsymbol{A}^\top\boldsymbol{A} = \boldsymbol{I}_d} \|\boldsymbol{X} - \boldsymbol{A}\boldsymbol{Z}\|_F^2 = \boldsymbol{U}\boldsymbol{V}^\top,
\end{equation}
where $\boldsymbol{X}\boldsymbol{Z}^\top = \boldsymbol{U}\boldsymbol{\Sigma}\boldsymbol{V}^\top$ is the compact SVD. Procrustes matches directions but not magnitudes, so we set the scale by matching Frobenius norms,
\begin{equation}
  c = \|\boldsymbol{A}^\top\boldsymbol{X}\|_F \,/\, \|\boldsymbol{Z}\|_F,
\end{equation}
which gives $c = 2.85$ for Wan2.1-T2V-14B and $c = 3.78$ for Wan2.1-T2V-1.3B.

\paragraph{Patch Embedding.}
Since $\boldsymbol{z} \approx \boldsymbol{A}^\top\boldsymbol{x} / c$, we initialize the pixel patch embedding as $\boldsymbol{W}\boldsymbol{A}^\top$, with the bias copied from the pretrained embedding. The $1/c$ factor is absorbed into the input calibration below. Components of $\boldsymbol{x}$ outside the span of $\boldsymbol{A}$ are initially discarded, and the patch embedding is trained with the rest of the model to pick them up.

\paragraph{Internal Timestep Calibration.}
Projecting a noisy pixel patch $\boldsymbol{x}_\sigma = (1-\sigma)\,\boldsymbol{x} + \sigma\,\boldsymbol{\epsilon}$ onto the lift gives
\begin{equation}
  \boldsymbol{A}^\top\boldsymbol{x}_\sigma \approx c(1-\sigma)\,\boldsymbol{z} + \sigma\,\boldsymbol{\epsilon}_z,
  \label{eq:proj_noisy}
\end{equation}
where $\boldsymbol{\epsilon}_z = \boldsymbol{A}^\top\boldsymbol{\epsilon} \sim \mathcal{N}(\boldsymbol{0}, \boldsymbol{I}_d)$ stays standard Gaussian since $\boldsymbol{A}$ has orthonormal columns. Its signal-to-noise ratio is that of the latent flow at the noise level $\sigma'$ satisfying $(1-\sigma')/\sigma' = c(1-\sigma)/\sigma$, \ie,
\begin{equation}
  \sigma' = \frac{\sigma}{c(1-\sigma) + \sigma}.
  \label{eq:time_calib}
\end{equation}
Rescaling the input by $k(\sigma) = \sigma'/\sigma = 1/\big(c(1-\sigma) + \sigma\big)$ further matches the magnitude,
\begin{equation}
  k(\sigma)\,\boldsymbol{A}^\top\boldsymbol{x}_\sigma \approx (1-\sigma')\,\boldsymbol{z} + \sigma'\,\boldsymbol{\epsilon}_z,
\end{equation}
which is a sample of the pretrained latent flow at noise level $\sigma'$. We therefore feed the DiT the input $k(\sigma)\,\boldsymbol{x}_\sigma$ with timestep $\sigma'$ throughout training and sampling. Since $c > 1$, $\sigma' < \sigma$: at the same $\sigma$, the redundant pixels carry more signal than the latents, and the calibration accounts for it by shifting the noise schedule seen by the DiT, as ablated in \cref{sec:ablation}. Unlike AsymFlow, which predicts velocity and inherits the pretrained output head through a calibrated output conversion, \methodname\ predicts $\hat{\boldsymbol{x}}$ with a newly initialized detailer, so no output calibration is needed.

\subsection{Detailer Head}
\label{sec:detailer}
\begin{figure*}[!t]
  \centering
  \includegraphics[width=\textwidth]{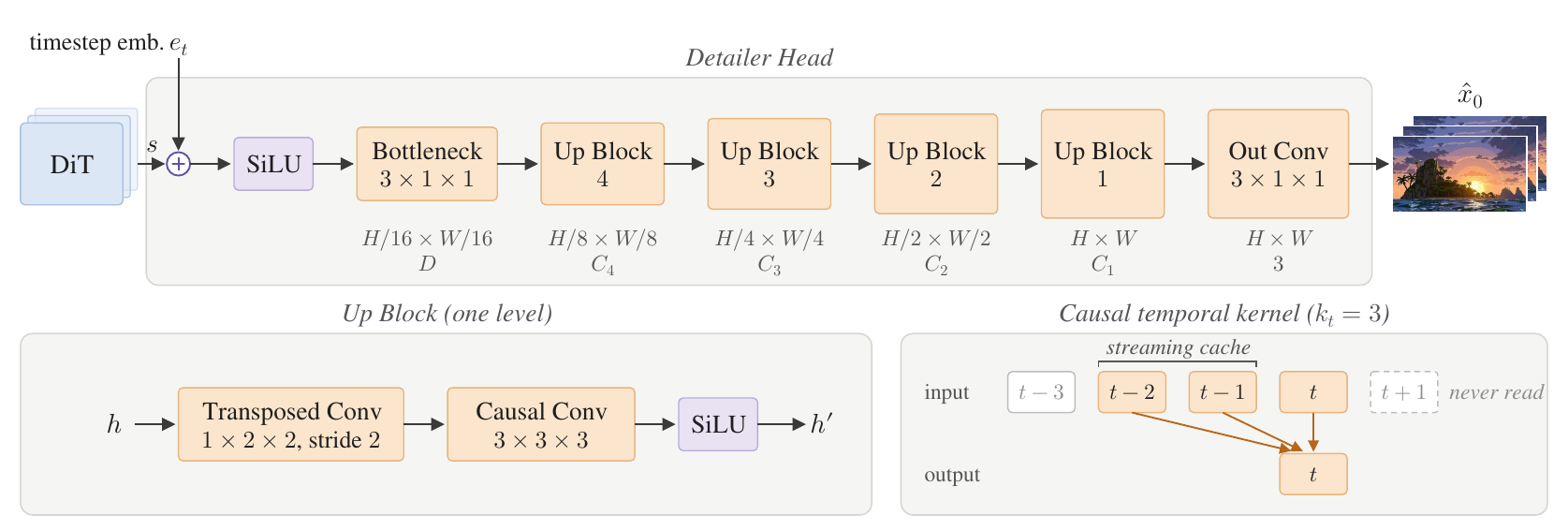}
  \caption{\textbf{Detailer head architecture} (patch size $p{=}16$).
    \emph{Top:} the DiT's pre-head tokens $s$, combined with the timestep embedding as
    $\mathrm{SiLU}(e_t + s)$, form a feature map on the token grid. A bottleneck convolution and
    $L{=}\log_2 p{=}4$ up blocks upsample it $2{\times}$ per level to full resolution, and an output
    convolution predicts $\hat{x}_0$. Feature sizes are per frame, where $D$ is the DiT width and
    $C_\ell$ the width of up block $\ell$.
    \emph{Bottom left:} each up block is a $2{\times}$ transposed convolution followed by a causal
    $3{\times}3{\times}3$ convolution and SiLU.
    \emph{Bottom right:} all convolutions are causal in time ($k_t{=}3$): output frame $t$ only reads
    frames $t{-}2$ to $t$, so in frame-by-frame generation the head keeps the previous two frames as a
    streaming cache.}
  \label{fig:detailer}
\end{figure*}

The detailer head replaces the linear output head and the VAE decoder of the pretrained latent model, mapping DiT features directly to pixels (Fig.~\ref{fig:detailer}). Its input is the last DiT hidden state $\boldsymbol{s}$ on the token grid of size $H/p \times W/p$ with width $D$. We add the timestep embedding $\boldsymbol{e}_t$ to $\boldsymbol{s}$, apply a SiLU, and project the result with a $3\times1\times1$ bottleneck convolution. A stack of Up Blocks then upsamples the features to the full resolution $H\times W$, and a final $3\times1\times1$ convolution outputs the pixel prediction $\hat{\boldsymbol{x}}$.
Since the head upsamples only spatially, it keeps the temporal resolution of the token grid.
Each Up Block doubles the spatial resolution, so a $1\times16\times16$ patch requires $4$ Up Blocks, with widths $C_4,\dots,C_1$.
We use $D=1536$ and $C=[512,256,128,64]$ for the 1.3B model, and $D=5120$ and $C=[1024,256,64,64]$ for the 14B model, giving detailer heads with $44$M and $337$M parameters, respectively.

\paragraph{Up Block.}
Each Up Block applies a $1\times2\times2$ transposed convolution to double the height and width, followed by a $3\times3\times3$ causal convolution and a SiLU. The transposed convolution upsamples each frame independently, and the causal convolution mixes information spatially and across neighboring frames at the new resolution.

\paragraph{Causal Decoding.}
All temporal kernels in the head have size $3$ and are padded only on the past side, so at every layer output frame $t$ depends only on input frames $t-2$, $t-1$, and $t$ (Fig.~\ref{fig:detailer}, bottom right). Therefore, the head never reads future frames and is compatible with causal, frame-by-frame generation. When decoding a stream, each convolution caches its two most recent input frames, so every new frame is decoded at a constant cost without recomputing earlier frames.

\subsection{Detailer Head for Higher Resolution}
\label{sec:detailer_highres}

To generate $960\times1664$ videos, we double the patch size from $1\times p\times p$ to $1\times2p\times2p$, so the token grid stays at $30\times52$ per frame while the resolution doubles. Table~\ref{tab:detailer_res} compares the two detailer heads.

\paragraph{Patch Embedding.} The $1\times2p\times2p$ patch embedding is built from the pretrained $1\times p\times p$ one by repeating each weight over a $2\times2$ block and scaling by $\frac{1}{4}$. This is exactly a $2\times2$ average pooling $\mathcal{P}$ followed by the pretrained patch embedding, so at initialization the DiT sees a $480$p-equivalent input.

\paragraph{Timestep Calibration.}\label{app:time_map}
Pooling keeps the signal but halves the standard deviation of the i.i.d.\ noise, so $2\,\mathcal{P}(\boldsymbol{x}_\sigma) \overset{d}{=} 2(1-\sigma)\,\mathcal{P}(\boldsymbol{x}) + \sigma\,\boldsymbol{\epsilon}$. This has the form of Eq.~\eqref{eq:proj_noisy} with $c=2$, so we apply the calibration of App.~\ref{sec:patch_embed_init} with scale $2$, \ie, timestep $\sigma/(2-\sigma)$ and input $\frac{2}{2-\sigma}\,\mathcal{P}(\boldsymbol{x}_\sigma)$, on top of the calibration of the patch-$p$ model. Composing two calibrations multiplies their scales, so the DiT sees a single calibration of Eq.~\eqref{eq:time_calib} with scale $2c$. We also double the noise shift, which keeps the effective shift $s/c$ unchanged, so the DiT sees the same noise schedule as the patch-$p$ model.

\begin{table}[t]
\centering
\footnotesize
\begin{tabular}{@{}lcc@{}}
\toprule
 & $480$p & $960$p \\
\midrule
Patch size $p$       & $1\times16\times16$ & $1\times32\times32$ \\
Tokens per frame     & $30\times52$ & $30\times52$ \\
Up blocks $L$        & $4$ & $5$ \\
Widths $C_L,\dots,C_1$ & $1024,256,64,64$ & $4096,1024,256,64,64$ \\
Head parameters      & $337$M & $942$M \\
Timestep shift       & $11$ & $22$ \\
\bottomrule
\end{tabular}
\caption{\textbf{Detailer head at $480$p vs.\ $960$p.}}
\label{tab:detailer_res}
\end{table}

\section{Experiment Details}
\label{sec:app_exp}

\subsection{Common Settings}
\label{sec:app_training}

Videos are represented in the OKLab color space~\cite{ottosson2020oklab}, with each channel normalized by its mean and standard deviation over the training data.
Unless otherwise noted, models are trained with a global batch size of $64$ on $64$ NVIDIA B200 GPUs.
 
We use AdamW~\cite{loshchilov2019decoupled} with $\beta_1=0.9$, $\beta_2=0.95$, and no weight decay, with a constant learning rate of $10^{-5}$ for pretrained or warm-initialized modules and $10^{-4}$ for randomly initialized ones, without warmup.
Training uses bfloat16, gradient clipping at $1.0$, and an exponential moving average (EMA) of the weights with decay $0.999$.
We set $\sigma_{\min}=0.05$, $\delta=10^{-3}$, and $\lambda_{\text{temp}}=\lambda_{\text{LPIPS}}=0.2$.
Both auxiliary losses are computed in sRGB on a random $256\times256$ crop, with LPIPS using a VGG backbone on $4$ random frames, and are weighted by $\omega(\sigma)/\max(\sigma,\sigma_{\min})^2$ with $\omega(\sigma)=(1-\sigma)^2/\big((1-\sigma)^2+\kappa^2\sigma^2\big)$ and $\kappa=0.3$.
We drop the text prompt with probability $0.1$.
At inference, we clamp the guided $\hat{\bm{x}}$ to the valid sRGB gamut at every step, and for text-to-video generation use classifier-free guidance~\cite{ho2022classifier} of scale $4.0$ with the default negative prompt of Wan2.1~\cite{wan2025wan}.
We do not clamp in precise video editing (App.~\ref{sec:app_d2v}), finding this produces more consistent colors.

\subsection{Text-to-Video Generation}
\label{sec:app_t2v}

We train on $64$K landscape videos generated by Wan2.1 at $720\times1280$, each with $81$ frames at $16$ fps and paired with its generation prompt, sampling random $21$-frame windows resized and cropped to $480\times832$.
Training takes about $57$ hours, and the model totals $14.6$B parameters (App.~\ref{sec:detailer}).

At evaluation, Wan2.1-T2V-14B generates $81$ frames at $16$\,fps with its official settings ($50$ UniPC steps, guidance $5.0$, shift $5$).
PixelUMM generates $96$ frames at $24$\,fps with its released settings ($35$ UniPC steps, guidance $6.0$, shift $10$).
Our models generate $21$ frames at $16$\,fps ($480\times832$ or $960\times1664$) with $50$ Euler steps, guidance $4.0$, and their training shifts ($11$ and $22$).
\Cref{tab:vbench} scores the first $1.3$\,s of every baseline video, first $21$ frames for Wan2.1 and first $31$ frames for PixelUMM.
\Cref{tab:vbench_full} repeats \cref{tab:vbench} with the baselines also scored over their full videos, $81$ frames for Wan2.1 and $96$ frames for PixelUMM.
Their total scores change by at most $0.10$ ($84.07$ and $83.82$), so scoring the first $1.3$\,s does not favor \methodname.

\paragraph{Human evaluation.}
We sample $100$ random prompts from Movie Gen Video Bench~\cite{polyak2024movie} and generate one video per model per prompt, showing raters $21$-frame, $1.3$\,s segments.
Each pair is shown side by side in random left/right order without model names, and raters choose the better video, or a tie, for three questions: ``Visual quality: Which video looks better in terms of sharpness, detail, and absence of artifacts?'', ``Motion quality: Which video has more natural and smooth motion, without flickering or objects deforming?'', and ``Text alignment: Which video better matches the text prompt?''.
$10$ raters provided $100$ votes each per comparison.

\begin{table*}[h]
\centering
\setlength{\tabcolsep}{2.2pt}
\resizebox{\textwidth}{!}{%
\begin{tabular}{@{}l c ccc ccccccc ccccccccc@{}}
\toprule
& & \multicolumn{3}{c}{Overall}
  & \multicolumn{7}{c}{Quality dimensions}
  & \multicolumn{9}{c}{Semantic dimensions} \\
\cmidrule(lr){3-5} \cmidrule(lr){6-12} \cmidrule(l){13-21}
Model & Size
  & \vbhead{Total} & \vbhead{Quality} & \vbhead{Semantic}
  & \vbhead{Subject\\consist.} & \vbhead{Backgr.\\consist.} & \vbhead{Temporal\\flicker.} & \vbhead{Motion\\smooth.} & \vbhead{Dynamic\\degree} & \vbhead{Aesthetic\\quality} & \vbhead{Imaging\\quality}
  & \vbhead{Object\\class} & \vbhead{Multiple\\objects} & \vbhead{Human\\action} & \vbhead{Color} & \vbhead{Spatial\\relation.} & \vbhead{Scene} & \vbhead{Appear.\\style} & \vbhead{Temporal\\style} & \vbhead{Overall\\consist.} \\
\midrule
Wan2.1-T2V~\cite{wan2025wan}, first 21 frames & 14B
  & 84.05 & 84.91 & 80.63
  & \vbsecond{97.22} & \vbsecond{98.20} & 99.21 & 98.28 & 60.83 & 66.60 & 67.77
  & 95.24 & 86.65 & 97.00 & 84.73 & 78.42 & 52.18 & \vbsecond{22.52} & 24.57 & 26.89 \\
Wan2.1-T2V~\cite{wan2025wan}, all 81 frames & 14B
  & 84.07 & 84.84 & 81.01
  & 94.96 & 97.64 & 99.18 & 98.18 & \vbsecond{66.39} & 67.15 & 67.81
  & 95.27 & 86.69 & \vbbest{97.60} & 85.36 & 78.86 & 52.66 & 22.43 & 24.83 & 27.13 \\
PixelUMM~\cite{wei2026pixelumm}, first 31 frames & 8B MoT
  & 83.72 & 84.19 & 81.86
  & \vbbest{97.27} & 96.75 & 99.25 & \vbsecond{98.69} & 58.33 & 62.56 & 68.72
  & 96.79 & \vbsecond{89.31} & 96.20 & \vbsecond{90.24} & 77.38 & 55.35 & 21.86 & \vbsecond{24.84} & \vbsecond{27.23} \\
PixelUMM~\cite{wei2026pixelumm}, all 96 frames & 8B MoT
  & 83.82 & 84.21 & \vbbest{82.25}
  & 95.46 & 95.52 & \vbsecond{99.31} & \vbbest{98.71} & 61.94 & 64.38 & 68.82
  & \vbsecond{96.93} & \vbbest{89.33} & \vbsecond{97.20} & \vbbest{90.59} & 78.97 & 55.73 & 21.65 & \vbbest{24.98} & \vbbest{27.34} \\
\midrule
\methodname\ 480p & 14B
  & \vbsecond{84.78} & \vbsecond{85.44} & \vbsecond{82.15}
  & 96.65 & 97.30 & 98.94 & 96.92 & \vbbest{73.89} & \vbsecond{67.30} & \vbsecond{71.25}
  & 95.66 & 88.87 & 96.20 & 81.59 & \vbbest{84.74} & \vbbest{57.75} & \vbbest{23.23} & 24.32 & 26.91 \\
\methodname\ 960p & 14B
  & \vbbest{85.09} & \vbbest{86.19} & 80.68
  & 97.00 & \vbbest{98.51} & \vbbest{99.59} & 98.31 & 65.00 & \vbbest{67.49} & \vbbest{71.83}
  & \vbbest{97.59} & 88.16 & 96.40 & 74.54 & \vbsecond{84.14} & \vbsecond{56.22} & 21.91 & 24.46 & 26.58 \\
\bottomrule
\end{tabular}%
}
\caption{\textbf{VBench with full-length baseline videos.}
Rows marked ``first'' repeat \cref{tab:vbench}, which scores the first 1.3\,s of every video, the length of our videos.
Rows marked ``all'' score the same baseline videos over their full length: 81 frames at 16\,fps (5.1\,s) for Wan2.1 and 96 frames at 24\,fps (4.0\,s) for PixelUMM.
Over the full videos, the baselines' total scores change by at most 0.10 (84.07 for Wan2.1, 83.82 for PixelUMM) and stay below \methodname's: longer clips lower subject consistency but raise dynamic degree, and the two changes largely cancel.
Best and second best are marked over all six rows.
{\setlength{\fboxsep}{1pt}\colorbox{vbbestbg}{\textbf{Bold}}: best; \colorbox{vbsecondbg}{\underline{underlined}}: second best.}}
\label{tab:vbench_full}
\end{table*}

\subsection{Efficient High-Resolution Generation}
\label{sec:app_highres_exp}

Starting from Wan2.1-T2V-14B, we first train a patch-16 pixel model 
at $480\times832$ for $9$K iterations (shift $11$, $28$ hours), and then finetune it at $960\times1664$ with patch $32$ for $20$K iterations (shift $22$, $73$ hours).
The $39.8$K training videos are generated using the captions of our text-to-video data as prompts; we sample random $21$-frame windows at $24$ fps, cropped and downsampled to the training resolution.

\subsection{Precise Video Editing}
\label{sec:app_d2v}

\paragraph{Data and Training.}
The training data are $79$K stock videos ($74\%$ ground-level, $26\%$ aerial) with captions and Video Depth Anything~\cite{chen2025video} depth, as $21$-frame windows at $16$ fps and $480\times832$.
The latent baseline, VACE-1.3B~\cite{vace}, passes both video and depth through the Wan2.1 VAE.
For \methodname, we convert Wan2.1-T2V-1.3B to pixel space ($20$K iterations) and add VACE's control branch, without the LPIPS loss.
Both models are finetuned for $10$K iterations, taking about $8$ hours for \methodname\ and $5$ for the latent model, and are sampled with the same caption and seed per clip.
Per $21$-frame clip, \methodname's DiT processes $32{,}760$ tokens per pass against the latent model's $9{,}360$, or $7.0\times$ the FLOPs ($28.3$ vs.\ $4.1$ PFLOPs over $100$ passes, excluding the control branch).
The zero-shot Wan2.1-Fun-1.3B-Control and Wan2.2-Fun-5B-Control models use ZoeDepth input maps, and Cosmos-Transfer2.5 completed $962$ of the $1{,}000$ clips.

\paragraph{Setup Details.}
\methodname\ uses patch size $16$ and a four-level detailer; depth enters only through the control tokens.
Both models are trained with shift $5$ and a learning rate of $10^{-5}$, or $10^{-4}$ for \methodname's newly initialized control parameters.
\methodname\ samples with $50$ Euler steps and the latent model with Wan's default $50$-step UniPC, both at shift $5$.
On $60$ separate clips, we select the guidance scale (and, for \methodname, whether to clamp) by choosing the setting whose static-region sharpness is closest to the source among those within $3\%$ of the lowest flow error. This yields a scale of $3$ for the latent model and a scale of $2.5$ without the clamp for \methodname.

\paragraph{Evaluation.}
The $1{,}000$ test clips come from $1{,}000$ held-out videos. We split them into four bins of $250$ clips each based on the $95$th percentile of source optical flow, namely slow ($<10.1$ px/frame), moderate ($10.1$--$29.5$), medium ($29.5$--$64.6$), and fast ($\geq64.6$).
Flow is estimated with NeuFlow v2~\cite{zhang2024neuflow}.
Motion error is the end-point error between the flows of the generated and source videos, and motion kept is the ratio of their $95$th-percentile flow magnitudes.
Depth error is the AbsRel between Video Depth Anything estimates on the generated and source videos, aligned per clip in scale and shift.
Temporal LPIPS compares the perceptual change between consecutive frames of the generated and source videos.
We report medians over clips, with per-bin results in \cref{tab:app_d2v}, except for FVD and FVMD, which are set-level. FVMD uses the $962$ clips completed by all models, and FVD uses all clips each model completed.
\methodname\ has lower motion error than the finetuned latent model on $88\%$ of fast motion clips ($84\%$ over all clips).

\begin{table}[t]
  \centering
  \footnotesize
  \setlength{\tabcolsep}{4pt}
  \begin{tabular}{@{}lccccc@{}}
    \toprule
    & \multicolumn{4}{c}{Source motion} & \\
    \cmidrule(lr){2-5}
    Model & Slow & Moderate & Medium & Fast & All \\
    \midrule
    \multicolumn{6}{@{}l}{\emph{Motion error (px)}$\downarrow$} \\
    VACE-1.3B~\cite{vace}       & 0.6 & 3.4 & 9.1  & 30.4 & 5.6 \\
    VACE-14B~\cite{vace}        & 0.4 & 2.4 & 6.7  & 24.7 & 4.1 \\
    Wan2.1-Fun-1.3B~\cite{aigc_apps_VideoX_Fun_2026} & 0.5 & 2.8 & 7.6 & 27.3 & 4.6 \\
    Wan2.2-Fun-5B~\cite{aigc_apps_VideoX_Fun_2026}   & 0.8 & 3.8 & 10.2 & 32.4 & 5.9 \\
    Cosmos-2.5-2B~\cite{nvidia2025worldsimulationvideofoundation} & 0.4 & 2.5 & 7.1 & 24.6 & 4.3 \\
    \addlinespace[2pt]
    Latent (finetuned)       & 0.3 & 1.6 & 4.6 & 17.9 & 2.7 \\
    \methodname\ (finetuned) & \textbf{0.2} & \textbf{1.1} & \textbf{3.1} & \textbf{10.7} & \textbf{1.9} \\
    \midrule
    \multicolumn{6}{@{}l}{\emph{Motion kept (\%), target 100}} \\
    VACE-1.3B       & 96 & 90 & 81 & 32 & 84 \\
    VACE-14B        & 94 & 95 & 92 & 63 & 91 \\
    Wan2.1-Fun-1.3B & 98 & 94 & 91 & 56 & 91 \\
    Wan2.2-Fun-5B   & \textbf{99} & 91 & 86 & 46 & 88 \\
    Cosmos-2.5-2B   & 86 & 91 & 86 & 51 & 84 \\
    \addlinespace[2pt]
    Latent (finetuned)       & \textbf{99} & 98 & 95 & 83 & 96 \\
    \methodname\ (finetuned) & \textbf{99} & \textbf{100} & \textbf{99} & \textbf{98} & \textbf{99} \\
    \midrule
    \multicolumn{6}{@{}l}{\emph{Depth error (AbsRel)}$\downarrow$} \\
    VACE-1.3B       & .129 & .182 & .219 & .270 & .199 \\
    VACE-14B        & .129 & .175 & .220 & .268 & .189 \\
    Wan2.1-Fun-1.3B & .166 & .217 & .237 & .275 & .221 \\
    Wan2.2-Fun-5B   & .243 & .274 & .308 & .344 & .288 \\
    Cosmos-2.5-2B   & .131 & .178 & .212 & .221 & .182 \\
    \addlinespace[2pt]
    Latent (finetuned)       & \textbf{.066} & \textbf{.099} & .119 & .136 & .103 \\
    \methodname\ (finetuned) & .074 & .101 & \textbf{.108} & \textbf{.109} & \textbf{.099} \\
    \midrule
    \multicolumn{6}{@{}l}{\emph{Temporal LPIPS ($\times100$)}$\downarrow$} \\
    VACE-1.3B       & 1.22 & 3.37 & 4.92 & 7.45 & 3.92 \\
    VACE-14B        & 1.03 & 2.44 & 3.75 & 5.50 & 2.91 \\
    Wan2.1-Fun-1.3B & 1.02 & 2.36 & 3.84 & 5.73 & 2.87 \\
    Wan2.2-Fun-5B   & 1.26 & 3.15 & 4.41 & 5.70 & 3.59 \\
    Cosmos-2.5-2B   & 1.19 & 3.70 & 4.45 & 4.81 & 3.54 \\
    \addlinespace[2pt]
    Latent (finetuned)       & \textbf{0.69} & \textbf{1.47} & 2.41 & 3.85 & 1.85 \\
    \methodname\ (finetuned) & 0.82 & 1.56 & \textbf{2.25} & \textbf{3.00} & \textbf{1.79} \\
    \bottomrule
  \end{tabular}
  \caption{\textbf{Precise video editing by source motion.}
    Medians over the 250 held-out clips in each motion bin (App.~\ref{sec:app_d2v}) and over all 1,000 clips; Cosmos-Transfer2.5 completed 962 clips.
    Motion kept is the generated video's flow magnitude as a percentage of the source's.
    The first five models are zero-shot; the last two are finetuned on our data with the same control branch.
    Set-level FVMD and FVD are in \cref{tab:d2v}.
    Bold: best per block.}
  \label{tab:app_d2v}
\end{table}

\paragraph{Error Analysis.}
On fast clips, the latent model's motion errors spike every four frames, at transitions between latents.
A VAE round trip of the source video has a motion error of $2.0$\,px on these clips and adds only $0.07$\,px at latent transitions, against $1.65$\,px for the latent model, so the errors arise in generation rather than in the VAE.

\subsection{Low-Latency Autoregressive Generation}
\label{sec:app_ar}

\paragraph{Setup.}
Both $4$-step students stream $81$ frames at $480\times864$ and $16$\,fps from a first frame, a caption, and a camera path.
SolarWM is the official SolarWM-5B student~\cite{huang2026solarwmopendatascalable}, which uses the Wan2.2 VAE with $4\times$ temporal compression.
For \methodname, we lift the bidirectional base with patch size $32$, matching its VAE's $16\times$ spatial compression and $2\times2$ DiT patch, and distill it with Self Forcing~\cite{huang2026self} for $3$K iterations with a noisy KV cache, keeping the $4$ sink frames clean.
Both students are compared with EMA weights.
As references, we also evaluate the bidirectional latent base at $50$ steps and Wan2.2-TI2V-5B~\cite{wan2025wan} without camera input.

\paragraph{Evaluation.}
The $807$ camera paths come from nine sources: ABot-World~\cite{jiang2026abot}, MultiCamVideo~\cite{bai2025recammaster}, DL3DV~\cite{ling2024dl3dv}, Sekai walking and game~\cite{li2025sekai}, RealCam-Vid~\cite{zheng2025realcam}, SpatialVID~\cite{wang2025spatialvid}, MiraData~\cite{ju2024miradata}, and OmniWorld~\cite{zhou2025omniworld}; SolarWM's camera-translation filter removed $93$ of $900$ planned clips.
FID uses $20$ frames per clip and FVD~\cite{unterthiner2018towards} $5$ windows of $16$ frames.
We also report the six VBench~\cite{huang2024vbench} dimensions that apply to arbitrary videos: subject and background consistency, motion smoothness, dynamic degree, aesthetic quality, and imaging quality (\cref{tab:app_ar_quality}).
Camera following uses poses estimated by ViPE~\cite{huang2025vipe}, reporting rotation error, translation error, CamMC, and frame-to-frame rotation error. 

\begin{table}[t]
  \centering
  \footnotesize
  \begin{tabular}{@{}lccc@{}}
    \toprule
    Metric & SolarWM~\cite{huang2026solarwmopendatascalable} & \methodname & \textcolor{gray}{Ground truth} \\
    \midrule
    FID$\downarrow$ & \textbf{12.65} & 12.70 & \textcolor{gray}{--} \\
    FVD$\downarrow$ & 53.8 & \textbf{52.4} & \textcolor{gray}{--} \\
    \midrule
    Subject consistency$\uparrow$    & 87.8 & \textbf{88.8} & \textcolor{gray}{91.2} \\
    Background consistency$\uparrow$ & \textbf{93.2} & 92.6 & \textcolor{gray}{94.4} \\
    Motion smoothness$\uparrow$      & 97.6 & \textbf{98.5} & \textcolor{gray}{98.2} \\
    Dynamic degree$\uparrow$         & 97.0 & \textbf{97.7} & \textcolor{gray}{93.2} \\
    Aesthetic quality$\uparrow$      & \textbf{56.1} & 54.1 & \textcolor{gray}{56.0} \\
    Imaging quality$\uparrow$        & \textbf{67.9} & 63.8 & \textcolor{gray}{68.7} \\
    \bottomrule
  \end{tabular}
  \caption{\textbf{Video quality for autoregressive generation} on the 807 camera paths (App.~\ref{sec:app_ar}).
    FID and FVD compare with the ground-truth videos; the VBench-style scores~\cite{huang2024vbench} are in percent, with ground-truth videos shown for reference.
    Bold: better of the two models.}
  \label{tab:app_ar_quality}
\end{table}

\paragraph{Responsiveness.}
The $100$ abrupt-turn paths ($50$ from RealCam-Vid, $50$ from DL3DV) are rotation-only and have no ground-truth video: each is a random sequence of holds, yaw turns and pitch turns at $1.5$--$2.5^\circ$ per frame, mixing $8$--$14$-frame moves with $2$--$5$-frame jerks.
We estimate the horizontal image shift between consecutive frames by phase correlation; Turn in \cref{tab:ar} is its correlation with the yaw control per frame, pooled over all clips, and \methodname's per-clip correlation is higher on every clip. Each model receives the whole path up front, so this measures how closely the motion follows the controls, not latency.
SolarWM receives the camera pose once per four-frame latent, at frames $0, 1, 5, \dots, 77$, so it cannot place a turn within a latent and spreads brief turns over neighboring frames, resulting in lower Turn of $0.70$ compared to \methodname\ at $0.95$.

\paragraph{Rollout Length.}
In \cref{tab:app_horizon}, we evaluate the same generations over the first $21$, $41$, and all $81$ frames.
\methodname's rotation error is slightly worse than SolarWM's over the first $21$ and $41$ frames, and its translation error is moderately higher, but the gap in camera following grows with the rollout length. In terms of visual quality, \methodname\ achieves lower FVD at every rollout length.
  \begin{table}[t]
    \centering
    \footnotesize
    \setlength{\tabcolsep}{2pt}
    \begin{tabular*}{\columnwidth}{@{\extracolsep{\fill}}llccc@{}}
      \toprule
      Rollout & Model & FVD$\downarrow$ & Rot.$\downarrow$ & Tr.$\downarrow$ \\
      \midrule
      \multirow{2}{*}{21 frames (1.3\,s)} & SolarWM~\cite{huang2026solarwmopendatascalable} & 58.2 & \textbf{1.53$^\circ$} & \textbf{0.255} \\
       & \methodname & \textbf{54.0} & 1.60$^\circ$ & 0.273 \\
      \midrule
      \multirow{2}{*}{41 frames (2.5\,s)} & SolarWM~\cite{huang2026solarwmopendatascalable} & 50.2 & \textbf{2.52$^\circ$} & \textbf{0.236} \\
       & \methodname & \textbf{46.1} & 2.76$^\circ$ & 0.268 \\
      \midrule
      \multirow{2}{*}{81 frames (5.1\,s)} & SolarWM~\cite{huang2026solarwmopendatascalable} & 53.8 & \textbf{4.24$^\circ$} & \textbf{0.231} \\
       & \methodname & \textbf{51.1} & 5.44$^\circ$ & 0.312 \\
      \bottomrule
    \end{tabular*}
    \caption{\textbf{Camera-controlled autoregressive generation across rollout lengths.}
    We computed over the first $21$, $41$, $81$ generated frames, following the same metrics as in \cref{tab:ar}.}
    \label{tab:app_horizon}
  \end{table}

\paragraph{Timing.}
We time both students on one B200 at $480\times864$ with $4$ denoising steps, after the local attention window has filled, on our custom inference stack built with optimized CuTeDSL kernels. A SolarWM latent covers $4$ pixel frames totalling $405$ tokens, while \methodname\ uses $405$ tokens for every frame, so SolarWM has higher throughput ($72.0$ vs.\ $29.4$\,fps).
For an isolated control, the control for frame $t$ arrives at $t/16$\,s and a block starts once its last control has arrived, so SolarWM's latency is its wait ($0$--$688$\,ms) plus its block compute time ($151$\,ms), giving $151$--$839$\,ms (median $495$\,ms), against $34$\,ms for \methodname.
SolarWM's official implementation, with its streaming VAE decoder, is slower: it takes $543$\,ms in the DiT and $205$\,ms in the VAE per $12$-frame block ($16.1$\,fps), for a median latency of $1{,}091$\,ms (\cref{tab:app_ar_cost}).
Per generated frame, \methodname's DiT needs $3.0\times$ SolarWM's FLOPs ($16.4$ vs.\ $5.6$ TFLOPs), and its throughput is $2.4\times$ lower.

\begin{table}[t]
  \centering
  \footnotesize
  \setlength{\tabcolsep}{2.6pt}
  \begin{tabular}{@{}llrrrrr@{}}
    \toprule
    Model & Code & Block & Tokens & TFLOPs & FPS & Latency \\
    \midrule
    SolarWM~\cite{huang2026solarwmopendatascalable} & official & 12 & 1,215 & 5.6 & 16.1 & 1,091\,ms \\
    SolarWM & ours & 12 & 1,215 & 5.6 & 72.0 & 495\,ms \\
    \methodname & ours & 1 & 405 & 16.4 & 29.4 & 34\,ms \\
    \bottomrule
  \end{tabular}
  \caption{\textbf{Cost of autoregressive generation} on one B200 at $480\times864$ with $4$ denoising steps.
    \methodname\ trades throughput for single-frame generation, cutting median control-to-screen latency from $495$\,ms to $34$\,ms while still streaming faster than real time.}
  \label{tab:app_ar_cost}
\end{table}

\paragraph{One-Latent SolarWM.}
\label{sec:app_1latent}
The official SolarWM student generates $3$-latent ($12$-frame) blocks, waiting up to $11$ frames for controls.
Since a latent model's minimum block is one latent ($4$ frames), we test whether shrinking SolarWM to one-latent blocks narrows the responsiveness gap to \methodname.
We train a $4$-step, one-latent SolarWM-5B student on the same \texttt{clean-81f} data ~\cite{huang2026solarwmopendatascalable} over a $15$-latent history window following SolarWM's official recipe, finetuning the teacher-forced AnyFlow Stage~1 checkpoint with one-latent blocks until convergence ($4$K iterations), followed by Stage~2 DMD distillation ($400$ iterations).

\begin{figure}[t]
  \centering
  \includegraphics[width=\columnwidth]{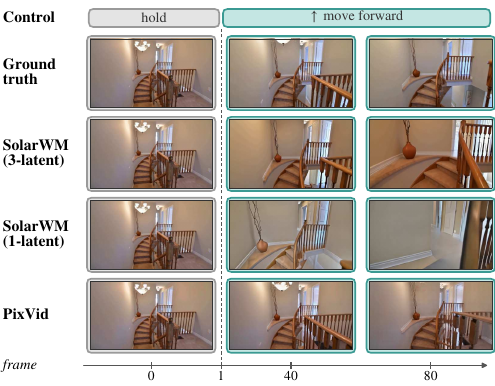}
  \caption{\textbf{Comparison with One-latent SolarWM.} SolarWM and PixVid show relatively small motion errors and produce moderately realistic videos. In contrast, One-latent SolarWM follows the controls less closely, here severely overestimating the motion, and looks less realistic.}
  \label{fig:app_1latent}
\end{figure}

\Cref{fig:app_1latent} shows One-latent SolarWM produces lower-quality camera following and output. This is reflected in \cref{tab:app_1latent}, which also shows latency of the one-latent model is still higher than \methodname.

\begin{table}[t]
  \centering
  \footnotesize
  \setlength{\tabcolsep}{1.5pt}
  \resizebox{\columnwidth}{!}{%
  \begin{tabular}{@{}lcccccccc@{}}
    \toprule
    & \multicolumn{4}{c}{Latency and speed} & Quality & \multicolumn{3}{c}{Camera} \\
    \cmidrule(lr){2-5}\cmidrule(lr){6-6}\cmidrule(lr){7-9}
    Model & Block & Wait$\downarrow$ & Meas.$\downarrow$ & FPS$\uparrow$ & FVD$\downarrow$ & Rot.$\downarrow$ & Tr.$\downarrow$ & Turn$\uparrow$ \\
    \midrule
    SolarWM~\cite{huang2026solarwmopendatascalable} & 12 & 11 & 495\,ms & \textbf{72.0} & 53.8 & \textbf{4.24$^\circ$} & \textbf{0.231} & 0.70 \\
    SolarWM (1-latent) & 4 & 3 & 164\,ms & 50.4 & 118.6 & 25.58$^\circ$ & 0.410 & 0.11 \\
    \methodname & \textbf{1} & \textbf{0} & \textbf{34\,ms} & 29.4 & \textbf{51.1} & 5.44$^\circ$ & 0.312 & \textbf{0.95} \\
    \bottomrule
  \end{tabular}%
  }
  \caption{\textbf{One-latent ($4$-frame) SolarWM ablations} compared with $12$-frame SolarWM and frame-wise \methodname\ on all metrics of \cref{tab:ar}.}
  \label{tab:app_1latent}
\end{table}

\subsection{Ablations}
\label{sec:app_abl}

All ablations use Wan2.1-T2V-1.3B with calibration scale $c=3.78$; the shift sweep uses the decoder-only head and $\bm{x}$-prediction, and the head and prediction ablations use $s=15$.
VBench in \cref{tab:abl_shift} uses all $946$ prompts with one video each, sampled and scored as in \cref{sec:t2v_exp}.
The $119$ prompts are every eighth of VBench's $946$, sampled with $50$ Euler steps in bfloat16.
All metrics are computed on grayscale frames.
Noise estimates the per-pixel noise standard deviation in 8-bit levels~\cite{immerkaer1996fast}.
Flicker applies the same estimator to the temporal residual, the video minus its temporal Gaussian blur ($\sigma=1.2$ frames).
Seam measures discontinuities on the $16\times16$ patch grid.
For each pixel column, we take the mean squared brightness difference to the next column over all rows and frames; at each patch border, we subtract the median of the six neighboring columns and take the median over borders, averaging the horizontal and vertical directions.
Noise that is independent across pixels adds the same amount at every column and cancels; we subtract the same quantity measured $4$ pixels from the patch borders, where no seam occurs, and report the result in squared 8-bit levels.

\subsection{Compute Cost}
\label{sec:app_cost}

\begin{figure}[t]
  \centering
  \includegraphics[width=\columnwidth]{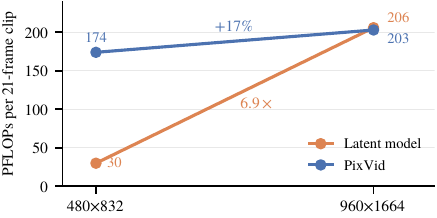}
  \caption{\textbf{Inference cost vs.\ resolution} for a 21-frame clip with the 14B DiT (100 passes; \cref{tab:app_t2v_cost}).
    Doubling the resolution gives a latent model $4\times$ the tokens and raises its FLOPs $6.9\times$, while \methodname\ keeps $30\times52$ tokens per frame, so only its detailer grows and its FLOPs rise by 17\%.}
  \label{fig:app_res_cost}
\end{figure}

Table~\ref{tab:app_t2v_cost} gives the cost of text-to-video inference.
FLOPs count the matrix multiplications of the DiT and the detailer head, including attention; we time each component with random weights on one B200, taking the median of $10$ runs after $3$ warmup runs.
Without temporal compression, a $21$-frame $480\times832$ clip has $3.5\times$ as many tokens as Wan2.1's latents ($32{,}760$ vs.\ $9{,}360$) and, since attention grows quadratically with the token count, $5.8\times$ the FLOPs.
Figure~\ref{fig:app_res_cost} shows doubling the resolution reverses this: a latent model's tokens grow $4\times$ and its FLOPs $6.9\times$, while \methodname\ keeps $30\times52$ tokens per frame, so only its detailer grows and its FLOPs rise by 17\%, matching the latent model at $960\times1664$.
At that resolution, \methodname\ still takes 9\% longer (192 vs.\ 176\,s), because its detailer runs convolutions on full-resolution frames, which use the GPU less efficiently than the DiT's matrix multiplications.

\begin{table}[t]
  \centering
  \footnotesize
  \setlength{\tabcolsep}{2.2pt}
  \begin{tabular}{@{}lcrrrrr@{}}
    \toprule
    Model & Frames & Tokens & Tok./s & PFLOPs & Time\,(s) & Mem.\,(GB) \\
    \midrule
    \multicolumn{7}{@{}l}{\emph{$480\times832$}} \\
    Wan2.1~\cite{wan2025wan} & 81 & 32,760 & 6,471 & 168 & 142 & 31.0 \\
    Wan2.1 & 21 & 9,360 & 7,131 & 30 & 26 & 30.8 \\
    \methodname & 21 & 32,760 & 24,960 & 174 & 151 & 33.4 \\
    \midrule
    \multicolumn{7}{@{}l}{\emph{$960\times1664$}} \\
    Latent$^\dagger$ & 21 & 37,440 & 28,526 & 206 & 176 & 41.9 \\
    \methodname & 21 & 32,760 & 24,960 & 203 & 192 & 53.6 \\
    \bottomrule
  \end{tabular}
  \caption{\textbf{Text-to-video inference cost} of the 14B models per clip on one B200, with $50$ sampling steps and classifier-free guidance.
    While removing temporal compression increases per-clip cost at $480\times832$, doubling the patch size at $960\times1664$ keeps token count and runtime comparable to a Wan2.1-VAE latent model.}
  \label{tab:app_t2v_cost}
\end{table}

\section{Qualitative Results}
To demonstrate the advantages of our approach, we show additional qualitative comparisons.

\subsection{Text-to-Video Generation}

\begin{figure*}[t]
  \centering
  \includegraphics[width=\textwidth]{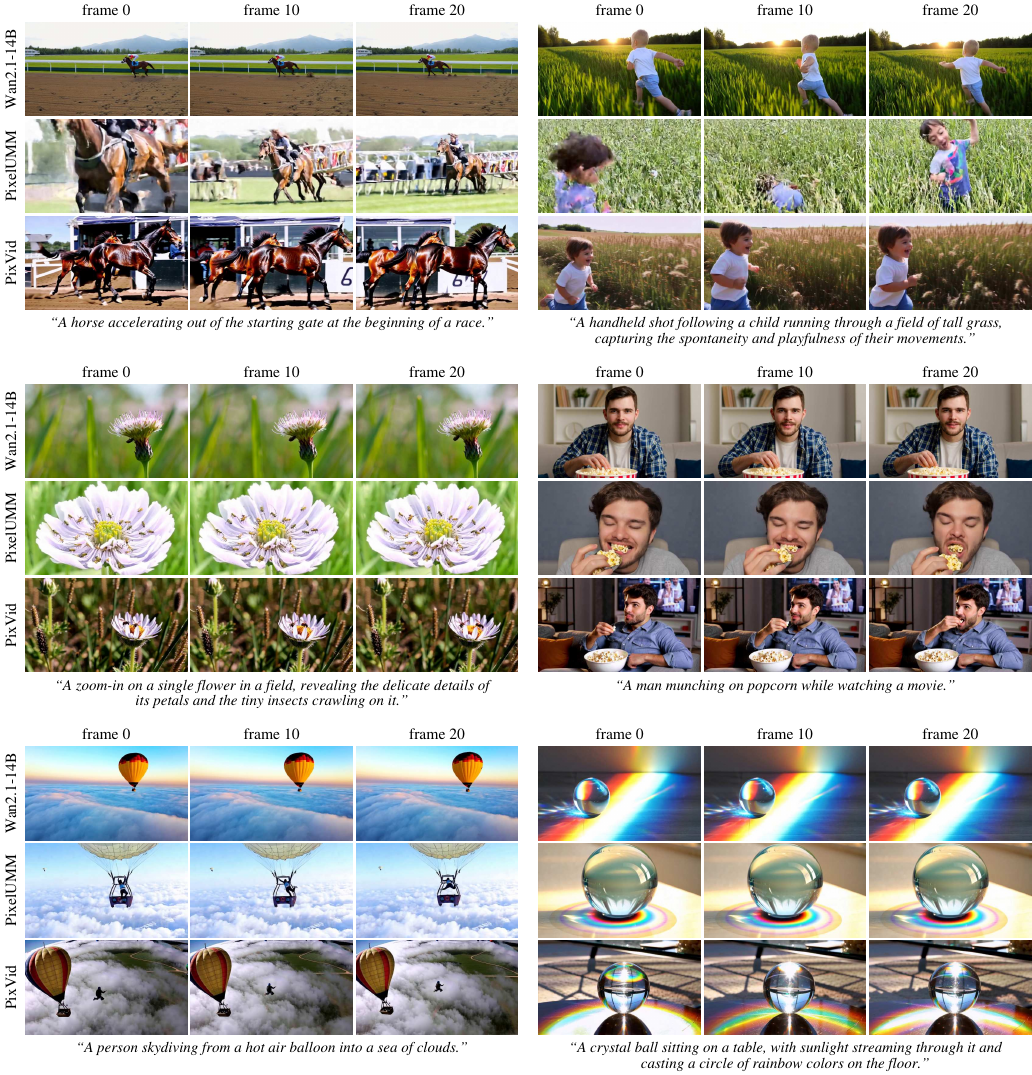}
  \caption{\textbf{Text-to-video comparison} on prompts from Movie Gen Video Bench~\cite{polyak2024movie}.
    We show frames 0, 10, and 20 of each 21-frame video from Wan2.1-T2V-14B, PixelUMM, and \methodname\ at $480\times832$.}
  \label{fig:t2v_qual}
\end{figure*}

In Fig.~\ref{fig:t2v_qual}, we show additional qualitative comparisons between \methodname, Wan2.1-T2V-14B and PixelUMM. \methodname produces high-quality videos, often with the most detail. PixelUMM struggles with more flicker and artifacts.

\subsection{High-Resolution Generation}

\begin{figure*}[t]
  \centering
  \includegraphics[width=\textwidth]{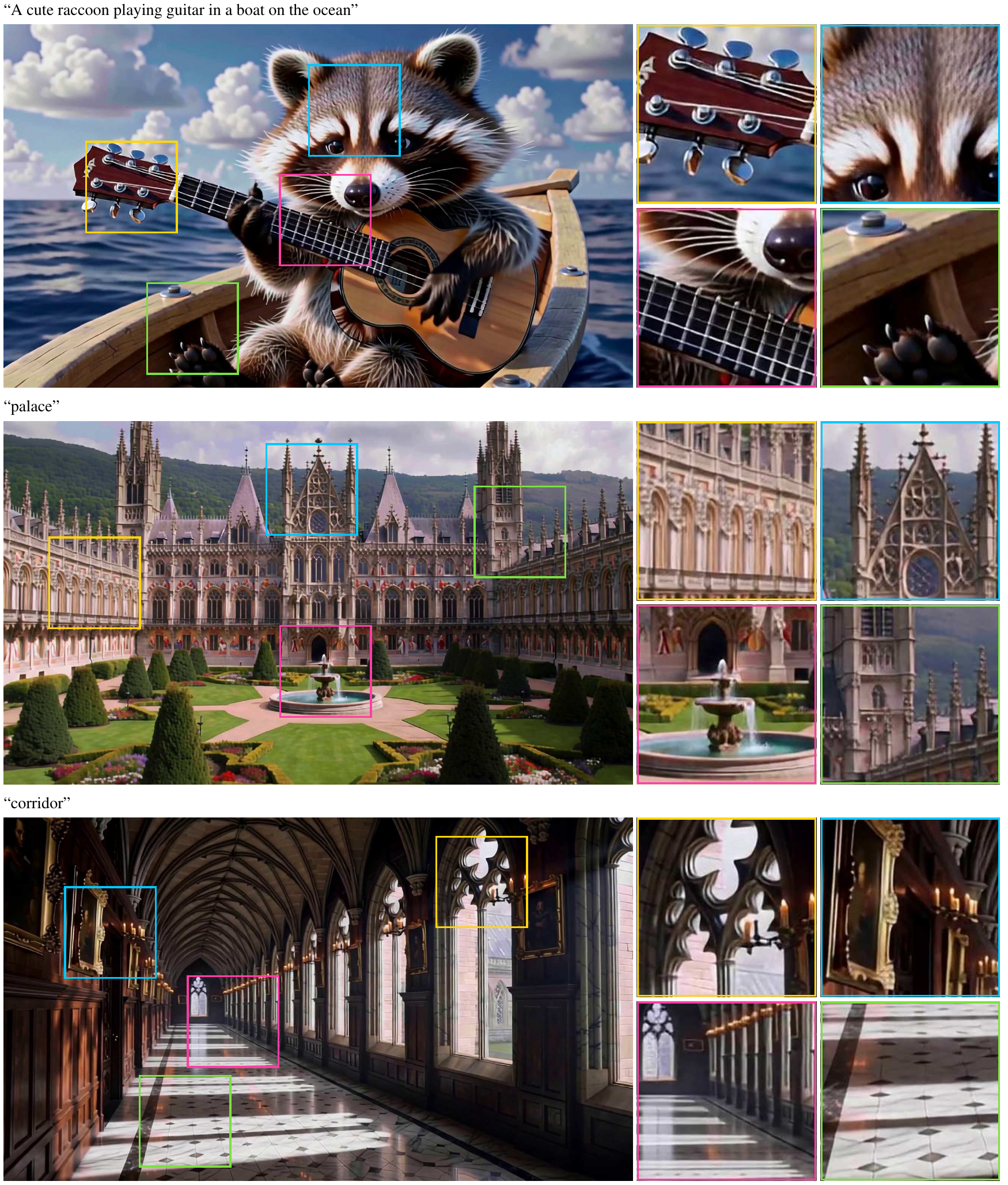}
  \caption{\textbf{High-Resolution Video Generation.} \emph{Left:} text-to-video samples     generated by our pixel model at a resolution of $1664\times960$. \emph{Right:} native-resolution crops of the boxed regions. Our method preserves fine pixel-level details such as fur, whiskers and architectural details because the model directly generates pixels without a VAE that compresses and reconstructs them.}
  \label{fig:app_high_res}
\end{figure*}

Initialized from Wan2.1-T2V-14B, which generates $81$-frame videos at $480\times832$, and trained as described in App.~\ref{sec:app_highres_exp}, \methodname spends the same token budget on resolution
instead of duration: our method generates $21$-frame videos at $960\times1664$ with much finer pixel-level details because it directly generates pixels without any VAEs that compress and reconstruct them, as shown in \cref{fig:app_high_res}. We show the \nth{10} frame of each generated video, and the full videos are on our project page.

\subsection{Precise Video Editing}

\begin{figure*}[t]
  \centering
  \includegraphics[width=\textwidth]{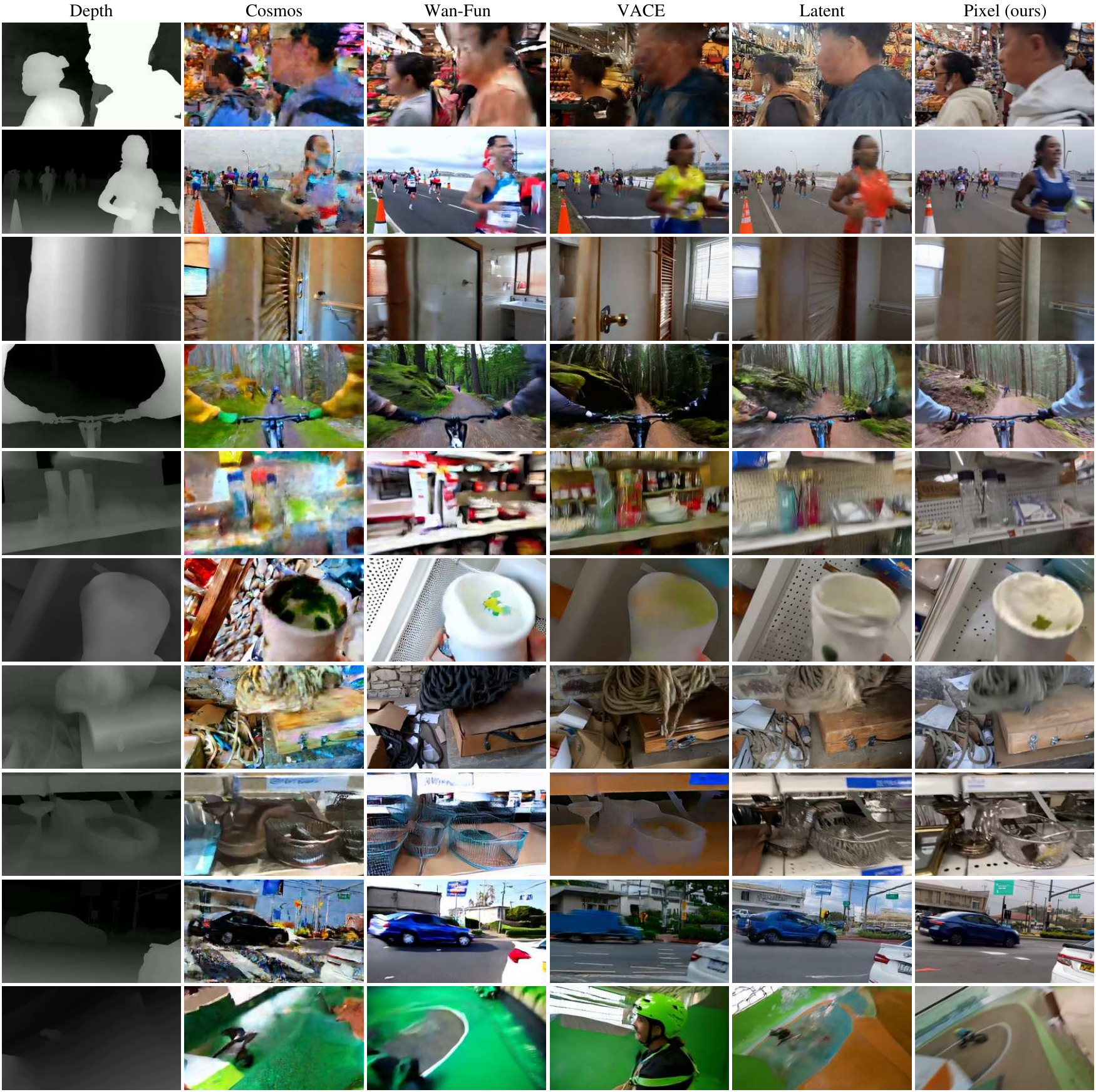}
  \caption{\textbf{Depth-to-video comparison on fast motion.} \methodname can generate high-quality outputs under sharp and fast motion.}
  \label{fig:depthcmp}
\end{figure*}

To evaluate how well our method handles sharp motion compared to latent-space approaches, we show additional depth-to-video results in \cref{fig:depthcmp}.
We compare \methodname-1.3B against Cosmos-Transfer2.5-2B~\cite{nvidia2025worldsimulationvideofoundation}, Wan2.1-Fun-1.3B-Control~\cite{aigc_apps_VideoX_Fun_2026}, VACE-1.3B~\cite{vace}, and VACE-1.3B finetuned with the same data and settings as ours.
We show one frame per clip from fast-motion clips; quantitative results over all $1{,}000$ held-out clips are in \cref{tab:app_d2v}.
Under such fast motion, the baselines, all of which compress several frames into each latent, often lose scene structure or blur, while our pixel-space model follows the depth more closely.

\subsection{Camera-Controlled Autoregressive Generation}

\begin{figure*}[t]
  \centering
  \includegraphics[width=\textwidth]{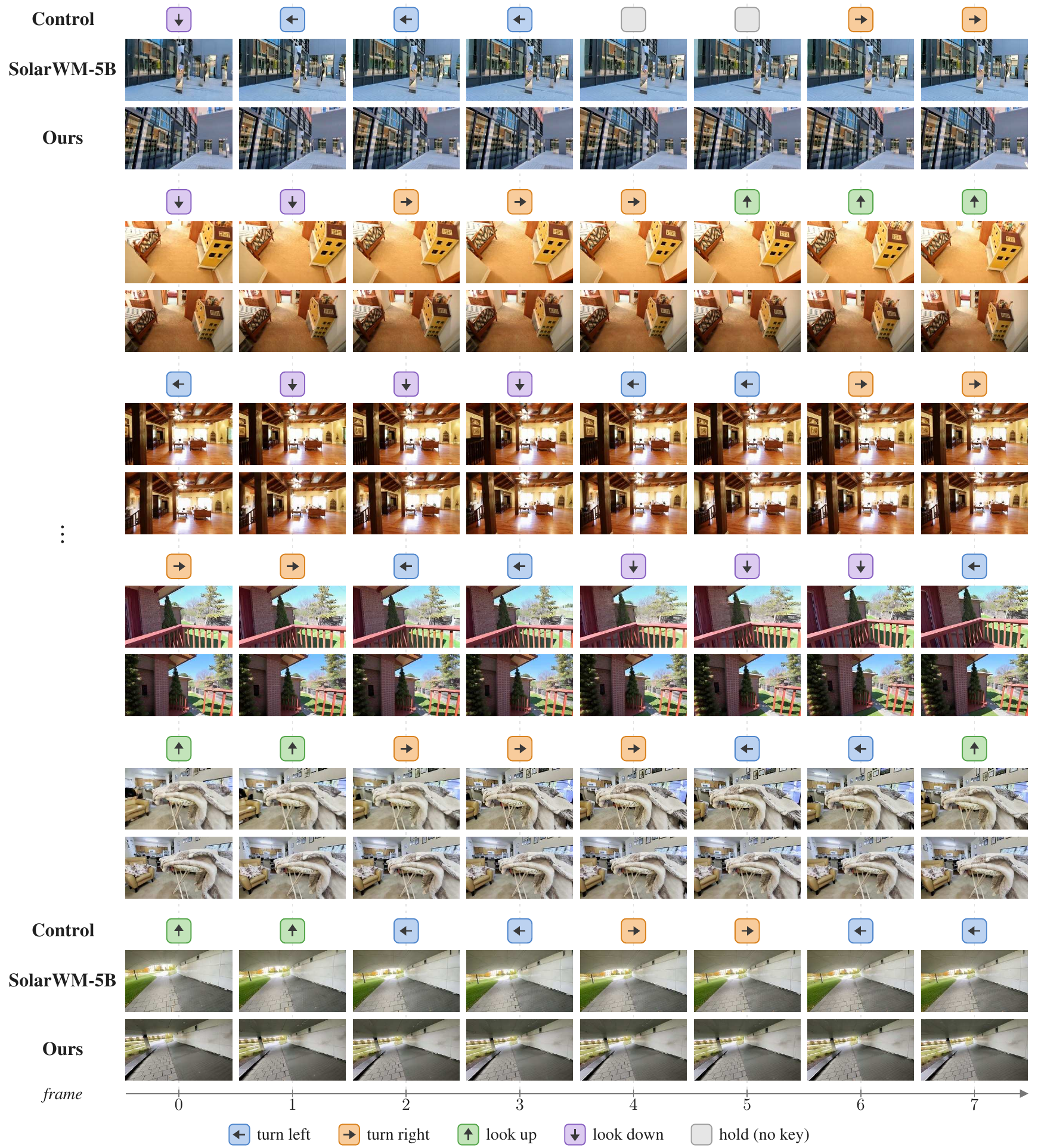}
  \caption{\textbf{Camera-controlled autoregressive generation.} We show $8$ consecutive generated frames at $16$~fps for simplicity.}
  \label{fig:camctrl_app}
\end{figure*}

In Fig.~\ref{fig:camctrl_app}, we show additional qualitative comparisons between SolarWM-5B and \methodname-5B under camera-controlled autoregressive generation. Both models use the same Wan~2.2 5B backbone and receive the same camera controls. 
\methodname-5B reaches a lower FVD (\cref{tab:ar}), although its imaging and aesthetic scores are lower (\cref{tab:app_ar_quality}). 
\methodname-5B accumulates drift during autoregressive generation, which leads to slightly larger rotation and translation errors. We leave this to future work.

\clearpage

\end{document}